%% file: iclr2027_conference.tex
\documentclass{article} 
\usepackage{iclr2027_conference,times}

\input{math_commands.tex}

\usepackage{hyperref}
\usepackage{url}
\usepackage[table]{xcolor}
\usepackage{graphicx}
\usepackage{fontawesome5}
\usepackage{booktabs}
\usepackage{makecell}
\usepackage{multirow}
\usepackage{tabularx}
\usepackage{caption}
\usepackage{enumitem}
\usepackage{array}
\usepackage[most]{tcolorbox}
\usepackage{listings}
\usepackage{needspace}
\usepackage{float}
\usepackage{xcolor}
\usepackage{pifont}

\definecolor{UGGreen}{HTML}{1B7837}
\definecolor{UGRed}{HTML}{B2182B}
\definecolor{UGGray}{HTML}{666666}

\newcommand{\UGYes}{\textcolor{UGGreen}{\ding{51}}}
\newcommand{\UGNo}{\textcolor{UGRed}{\ding{55}}}

\usepackage{titletoc}

\titlecontents{section}
  [0em]
  {\addvspace{0.25em}\small\bfseries}
  {\contentslabel{1.8em}}
  {}
  {\titlerule*[0.6pc]{.}\contentspage}
\titlecontents{subsection}
  [1.8em]
  {\small}
  {\contentslabel{2.7em}}
  {}
  {\titlerule*[0.6pc]{.}\contentspage}

\newcolumntype{B}{
    >{
        \columncolor{ModelHeadLight!70}[0pt][0pt]
        \centering
        \arraybackslash
    }m{1.30cm}
}

\newcolumntype{E}{
    >{
        \centering
        \arraybackslash
    }m{1.30cm}
}

\definecolor{LevelOne}{HTML}{4C78A8}
\definecolor{LevelOneLight}{HTML}{EAF1F8}

\definecolor{LevelTwo}{HTML}{2A9D8F}
\definecolor{LevelTwoLight}{HTML}{E7F4F2}

\definecolor{LevelThree}{HTML}{8B6FB1}
\definecolor{LevelThreeLight}{HTML}{F0ECF6}

\definecolor{LevelFour}{HTML}{D99A35}
\definecolor{LevelFourLight}{HTML}{FBF1DF}

\definecolor{LevelFive}{HTML}{C45D68}
\definecolor{LevelFiveLight}{HTML}{F8E8EA}

\definecolor{ModelHead}{HTML}{4B5563}
\definecolor{ModelHeadLight}{HTML}{F0F2F4}

\definecolor{Average}{HTML}{607D8B}
\definecolor{AverageLight}{HTML}{E8EEF1}

\DeclareRobustCommand{\Lone}[1]{\textcolor{LevelOne}{#1}}
\DeclareRobustCommand{\Ltwo}[1]{\textcolor{LevelTwo}{#1}}
\DeclareRobustCommand{\Lthree}[1]{\textcolor{LevelThree}{#1}}
\DeclareRobustCommand{\Lfour}[1]{\textcolor{LevelFour}{#1}}
\DeclareRobustCommand{\Lfive}[1]{\textcolor{LevelFive}{#1}}

\DeclareRobustCommand{\UrbanGroundName}{\texorpdfstring{\textsc{\Lone{U}\Ltwo{r}\Lthree{b}\Lfour{a}\Lfive{n}Ground}}{UrbanGround}}

\input{paper_revision.tex}

\input{urbanground_arxiv_authors}

\begin{document}

\maketitle
\UrbanGroundPostTitle
\UrbanGroundCorrespondenceNote

\UrbanGroundProjectLinks
\UrbanGroundHeroFigure

\begin{urbangroundabstract}
\input{sections/abstract}
\end{urbangroundabstract}


\input{sections/introduction}
\section{Related Work}
\label{sec: Related Work}

Spatial agency requires an agent to connect what is visible at the current moment to a spatial frame that remains usable after the view changes. 
Evaluation environments determine whether this transition from local competence to movement through a larger world can be observed. 
Existing work studies different parts of this problem through game environments and physical-world environments.

\paragraph{Agent Evaluation in Game Environments.}

Game environments provide a controlled setting for studying repeated perception and action, and early benchmarks use them to test whether multimodal models can turn visual observations into effective decisions~\citep{BALROG, V-MAGE, lmgame-Bench, VideoGameBench}. 
Other benchmarks isolate visual-spatial reasoning or make task completion verifiable through explicit environment state~\citep{iVISPAR, GameWorld}. 
These settings expose failures in visual control while keeping action consequences easy to measure. 

Persistent 3D worlds extend this evaluation across larger environments and longer trajectories~\citep{MineDojo, GROOT, JARVIS-1, MCU}. 
Later work further studies whether agents can preserve useful state across extended exploration and complete sustained objectives in game environments~\citep{ROCKET-1, Odyssey, MineExplorer, Orak, Cradle, Lumine, luo2026aigamesfoundationmodel}. 
These platforms establish that current agents remain fragile as interaction continues. 
However, their spatial structure is still defined by the game itself, so progress within a task does not directly show whether an agent has anchored its current observation to a spatial frame that remains valid beyond the game environments.

\paragraph{Agent Evaluation in Physical-World Environments.}

Physical-world evaluation places perception and action inside scene geometry. 
Indoor benchmarks study embodied question answering and navigation in bounded environments~\citep{OpenEQA, EmbodiedBench, NavBench, VLN-MME, CapNav}, while navigation models repeatedly convert visual history into local motion~\citep{NaVid, NaVILA, Embodied_Model, Qwen-RobotNav}. 
Open-vocabulary target search further tests whether an agent can locate a semantically specified destination under physical constraints~\citep{IndustryNav, VLNVerse}. 
These settings ground behavior in physical space, but their bounded extent limits how far a local observation must be related to a larger spatial frame.

Urban environments increase this spatial extent and make the relation between local observation and global location more consequential. 
Work grounded in real cities uses street-view imagery, aerial observations, or recorded trajectories to evaluate navigation at urban scale~\citep{V-IRL, VELMA, CityNavAgent, Exploring_Emergent_Navigation, CitySeeker, CityWalker, UrbanNav}. Interaction in these settings remains mediated by visual views, so changes in viewpoint do not arise from continuous physical movement through the same city geometry. 
Recently, interactive urban simulators restore closed-loop control~\citep{MetaUrban, UrbanWorld, SimWorld-Robotics, Virtual_Community}, while their environments are generated or assembled for simulation and do not preserve the full georegistered complex structure of an existing metropolis (see Appendix~\ref{app:urban_environment_comparison}). 
This leaves unresolved whether strong local multimodal competence can compose into spatial agency when observation and movement must remain consistent with the same environment.

\section{\UrbanGroundName{}: A Framework for Evaluating Spatial Agency}
\label{sec: Implementation of UrbanGround}
\label{app:framework-implementation}

\UrbanGroundName{} is a Unity-based framework that turns territory-scale geospatial data into a physically constrained city for evaluating MLLM agents.
It is designed to reveal whether evidence grounded at one viewpoint remains useful after the agent begins to move.

\subsection{Spatial Agency in Closed-Loop Urban Interaction}
\label{sec: spatial agency}

We define spatial agency through three cumulative capabilities.
\emph{Grounding} forms task-relevant local relations from locally acquired observations.
\emph{Persistence} keeps those relations usable after earlier evidence leaves the first-person view.
For global tasks, this also requires the model to align map context with the geometry encountered on the ground.
\emph{Adaptation} concerns whether the state remains reliable when external conditions alter visible evidence.
When new evidence invalidates an earlier assumption, adaptation requires a corresponding revision of the working state.
RQ1 primarily probes grounding.
RQ2 retains local grounding while stressing persistence over navigation.
RQ3 presupposes both capabilities and examines robustness under changes in the visible city.

Let \(s_t\) denote the hidden interaction state at turn \(t\).
It contains the georegistered agent pose and current interface state.
Task progress and city conditions are also included.
The evaluated model does not read \(s_t\) directly.
Its model-visible observation bundle is \(o_t\), which includes the current first-person or map view plus any task-specific notice or status exposed at that turn.
The model-facing interface exposes no privileged information, including remaining distance, shortest path, or route API.
We use \(g\) for the task objective, \(m_t\) for the working spatial state carried in the interaction context, and \(a_t\) for one structured action executed by the simulator.
The closed loop is
\begin{equation}
\begin{aligned}
o_t &= \mathcal{O}(s_t), \\
m_0 &= \mathcal{U}_0(o_0, g), \\
m_t &= \mathcal{U}(m_{t-1}, o_t, a_{t-1}, g), \quad t \geq 1, \\
a_t &= \pi(m_t, g), \\
s_{t+1} &= \mathcal{T}(s_t, a_t, \xi_t).
\end{aligned}
\label{eq:closed-loop-spatial-agency}
\end{equation}
The variable \(m_t\) is an analytical description of task-relevant spatial information induced in the model's interaction context over successive observation–action turns.
It does not assume an explicit memory module and is not observed by the evaluator.
For matched time and weather episodes, the scene condition varies in \(s_0\) while \(\xi_t\) remains fixed.
The term \(\xi_t\) can change during execution for a road-closure event or pedestrian motion.
Each executed non-terminal action updates the physical or interface state, which determines the next evidence \(o_{t+1}\).

The framework realizes this loop through three layers.
The \emph{geospatial layer} registers the city mesh and pedestrian network in a shared geographic frame.
The \emph{simulation layer} implements \(\mathcal{T}\) through continuous motion with collision and controlled changes in time, weather, and pedestrian activity.
The \emph{agent layer} exposes \(\mathcal{O}\) through first-person RGB observations and an interactive map without computed routes, and accepts structured physical and map actions.
This shared world and interface remain fixed across evaluated models.
The following subsections provide the implementation details. 
Within this common setting, our experiments progress from local environment understanding to navigation and multi-task planning, then to adaptation under changing conditions.

\subsection{Geospatial Layer}

\begin{figure*}[t!]
  \centering
  \includegraphics[width=0.98\textwidth]{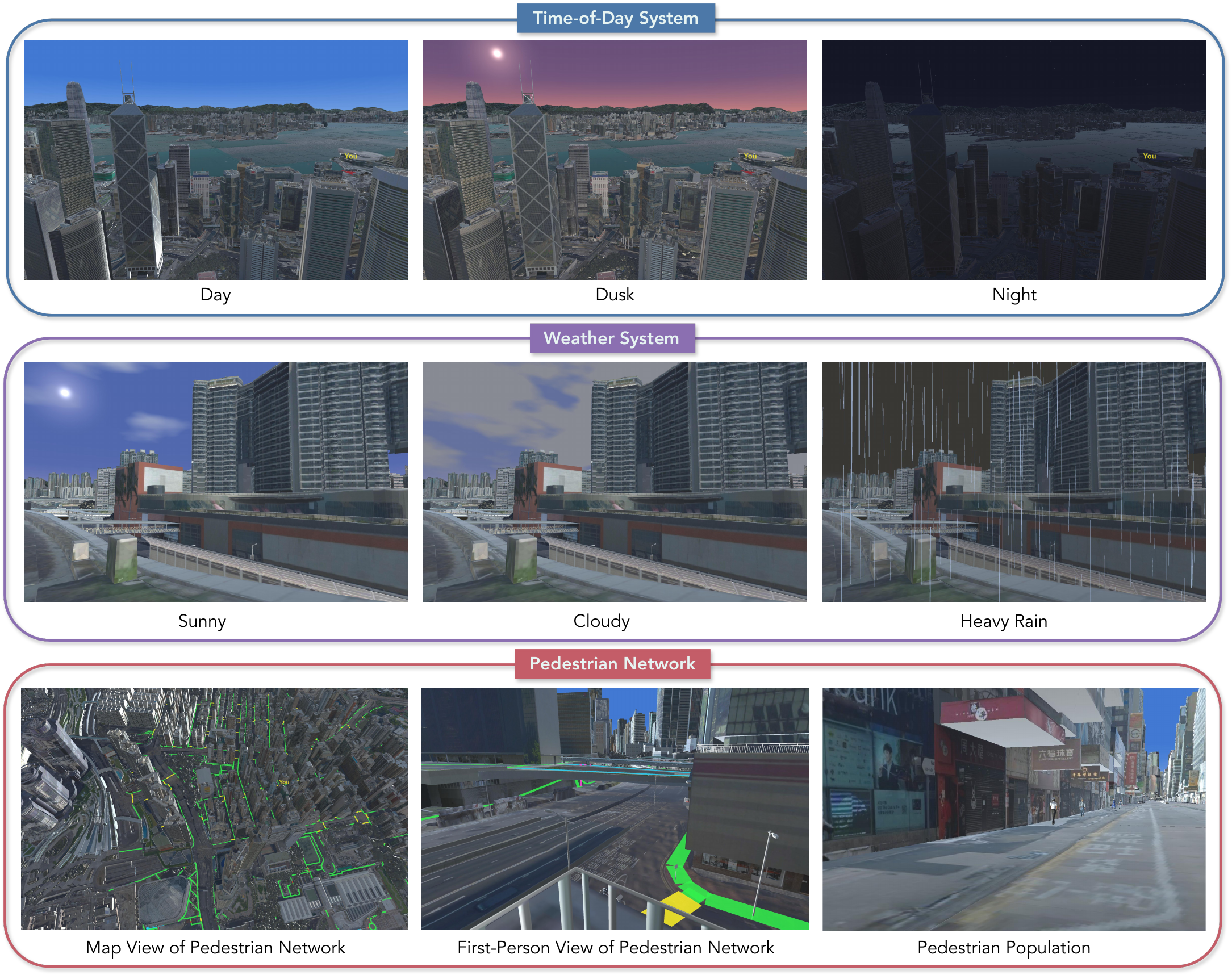}
  \caption{Dynamic simulation components of \UrbanGroundName{}. The same urban environment can be rendered under different times of day and weather conditions, while a georeferenced pedestrian network supports animated pedestrian populations.}
  \label{fig: scenarios}
\end{figure*}

\UrbanGroundName{} is built on the 3D Digital Map released by the Hong Kong Lands Department, which provides territory-wide representations of the city. 
We use two datasets from this collection. 
The \emph{3D Visualisation Map} provides a tile-based textured mesh reconstructed from oblique aerial imagery and covers the full territory of Hong Kong. 
It is distributed through an open API as Cesium 3D Tiles under the WGS84 reference system. 
The \emph{3D Pedestrian Network} provides georeferenced 3D line features derived from pedestrian-related road records. 
The former captures the visible form of the city, while the latter describes how its pedestrian spaces are connected.

\UrbanGroundName{} loads the 3D Tiles hierarchy into Unity at runtime and transforms the geographic coordinates of each tile into a shared simulation frame. 
Tiles are streamed according to the current viewing scale. 
The active tile geometry supplies both the rendered urban surface and the collision geometry used by the character controller. 
As the agent moves, the loader updates the active region while preserving its position in the global geographic frame.

The pedestrian network is imported into the same coordinate system as a connected graph. 
Each vertex retains its geographic position and associated attributes, while each edge represents a pedestrian segment between two connected locations. 
It does not restrict the controlled agent to predefined edges. 
The agent remains free to move in continuous space, while its trajectory can be compared with the registered network after execution.

\subsection{Simulation Layer}

The simulation layer turns the registered city into a continuously evolving environment. 
It controls embodied movement, collision, time, weather, and pedestrian activity while preserving the geographic coordinates supplied by the underlying data (Figure~\ref{fig: scenarios}).

\paragraph{Embodiment and physical constraints.}

The controlled agent is instantiated as a first-person character in Unity.
Its motion is continuous and resolved against collision geometry derived from the city mesh. 
Buildings, walls, and elevation changes therefore constrain the trajectory that the agent can execute. 
The agent moves on the surface shown in its observation and cannot cross visible structures through coordinate updates alone.

The 3D pedestrian network defines the intended walkable space but does not directly constrain the agent to a graph edge. 
The agent remains free to move in continuous space. 
These behaviors are recorded in the trajectory and later support analyses of off-network movement, route efficiency, and repeated exploration.

At every simulation step, \UrbanGroundName{} records the agent's metric position, elevation, and heading. These states support first-person control and provide the basis for determining arrival, traveled distance, orientation, and collision events.

\paragraph{Time-of-day system.}

\UrbanGroundName{} maintains a continuous simulation clock that controls the sky, sun position, ambient illumination, and shadows. 
Advancing the clock changes the appearance of the same location from day to dusk and night. 
Artificial lighting becomes active after dark, producing street-level observations whose visibility differs from daytime conditions.

The clock may be fixed at a specified hour for controlled comparisons or allowed to advance during an episode. 
Time is stored as part of the environment state and logged with each observation, allowing trajectories to be reproduced under the same temporal setting.

\paragraph{Weather system.}

\UrbanGroundName{} supports configurable weather conditions including rain and fog.
Weather alters visibility, illumination, atmospheric appearance, and surface rendering. Rain additionally produces particles that interact with the scene, which allows the simulator to determine whether the agent is exposed at its current position.

This exposure signal distinguishes covered paths from open streets along an executed trajectory. 
It supports tasks in which route quality depends on the physical conditions encountered during navigation, including whether an agent selects a sheltered path under rain.

\paragraph{Pedestrian population.}

A city is not empty, and neither is \UrbanGroundName{}. 
We populate the sidewalks with animated pedestrians drawn from the open-source Microsoft Rocketbox avatar library~\citep{Rocketbox}. 
Pedestrians are spawned on the registered network and assigned routes over its edges. 
Pedestrian positions and collisions are recorded at every step. 
Navigation success can therefore be considered together with the safety of the executed trajectory. 

\subsection{Agent Layer}
\label{sec: Agent Layer}

The agent layer connects an external MLLM to the running Unity environment. 
The simulator and the model operate as separate processes and communicate through a client-server interface. 
This allows different MLLMs and agent frameworks to control the same embodiment without modifying the Unity implementation. 
While awaiting a model response, the Unity simulation is paused. 
Simulation time advances only during simulator-side action execution, so differences in API latency do not affect the evaluated trajectories.

\paragraph{Observation space.}
The primary observation is an RGB image rendered from the agent's first-person camera. 
It presents the real city from the pose produced by the previous physical action. 
A bounded frame buffer may be included in the observation context to provide visual history across consecutive decisions.

For tasks that require global reasoning, \UrbanGroundName{} provides an interactive map tool. 
The map presents a georeferenced overhead view of the city and marks the agent's current location. 
The agent can pan and zoom the map to inspect areas outside its immediate first-person view.
The map does not expose a computed or highlighted route.

\paragraph{Action space.}
At each interaction turn, the agent selects exactly one action object from
\begin{equation}
\mathcal{A}
=
\mathcal{A}_{\mathrm{fp}}
\cup
\mathcal{A}_{\mathrm{map}}
\cup
\{\texttt{terminate}\},
\end{equation}
where
\begin{equation}
\mathcal{A}_{\mathrm{fp}}
=
\{
\texttt{move},
\texttt{sprint},
\texttt{look},
\texttt{jump},
\texttt{open\_map}
\}
\end{equation}
and
\begin{equation}
\mathcal{A}_{\mathrm{map}}
=
\{
\texttt{map\_select},
\texttt{map\_pan},
\texttt{map\_zoom},
\texttt{map\_orbit},
\texttt{close\_map}
\}.
\end{equation}

The \texttt{move} and \texttt{sprint} actions specify one of four movement directions and a duration.
They may also include \texttt{yaw\_rate}, \texttt{pitch\_rate}, \texttt{jump}, and \texttt{jump\_at} fields, allowing the agent to adjust its view or jump during the same movement action.
The direction field already represents lateral movement, so \texttt{strafe} is not a separate action.
The \texttt{look} action changes yaw and pitch without translation, while \texttt{jump} supports a standalone jump.
The map actions open and close the map, select a point for inspection, and control the map view through panning, zooming, and orbiting.
The \texttt{terminate} action ends the episode.

\subsection{Spatial Agency Evaluation Ladder}
\label{sec:spatial-agency-ladder}

\begin{figure*}[t!]
  \centering
  \includegraphics[width=0.98\textwidth]{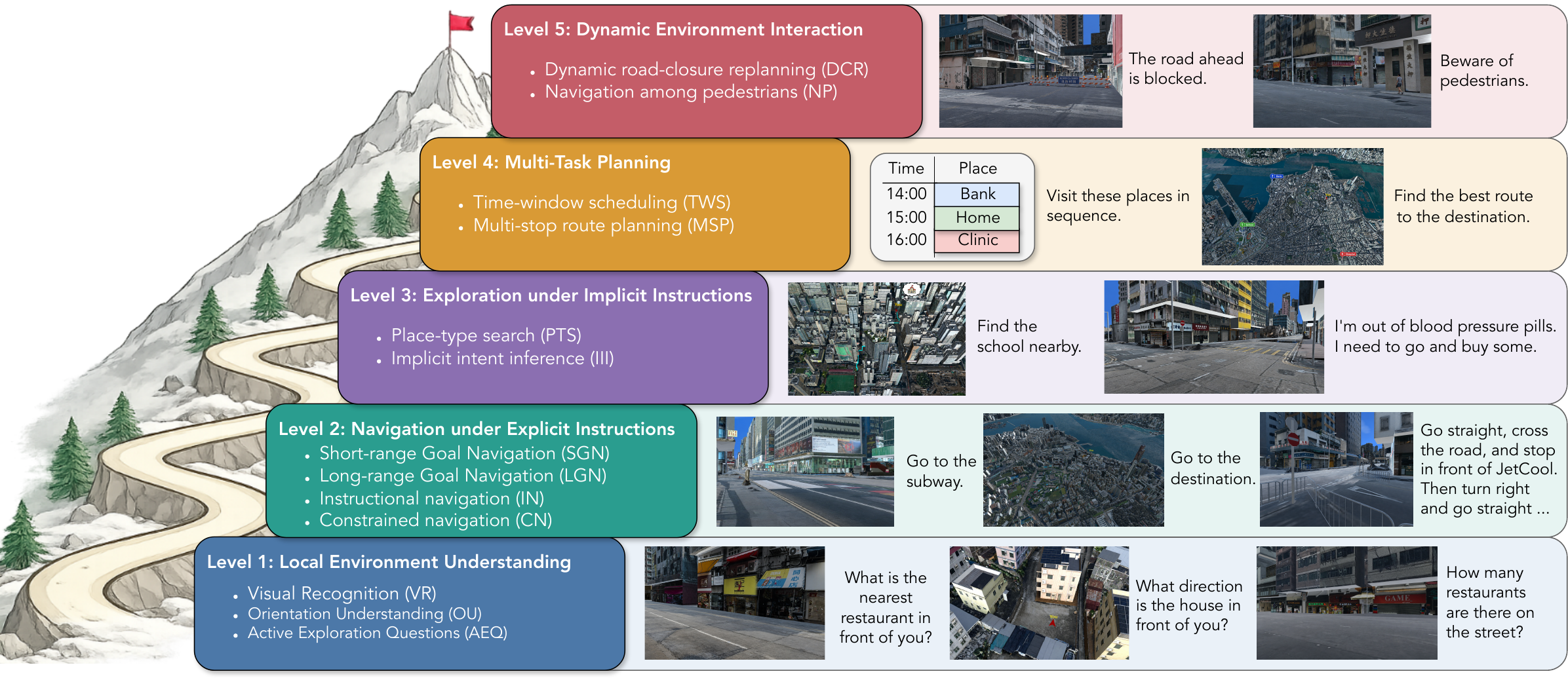}
  \caption{The spatial agency evaluation ladder increases the state that must remain usable across action. Level~1 supports RQ1. Levels~2--4 support RQ2. Level~5 and matched visual interventions support RQ3.}
  \label{fig:rq-level-map}
\end{figure*}

We organize the tasks as a five-level ladder that increases the spatial state required for success while keeping the interaction interface fixed (Figure~\ref{fig:rq-level-map}).
Level~1 studies local environment understanding.
Visual recognition and orientation test what can be grounded around the current pose.
Active exploration adds a brief search for evidence outside the initial view.
This level provides the main tasks for RQ1.

Level~2 introduces navigation under explicit instructions.
Short- and long-range tasks vary how long grounded evidence must remain useful.
Instructional and constrained variants test whether route language or path restrictions remain active during execution.
Level~3 moves to exploration under implicit instructions.
Place search and intent inference require the agent to infer where to go before it can navigate.
Level~4 extends persistence to multi-task planning.
The agent maintains and executes a plan across several goals under task-provided progress updates.
Levels~2--4 provide the tasks for RQ2 and stress persistence while retaining local grounding.

Level~5 introduces dynamic environment interaction.
A one-time system notice discloses a road closure that invalidates an earlier route.
The unavailable segment remains marked on the map.
Moving pedestrians require the agent to adjust its motion around visible obstacles.
Matched time and weather conditions apply to Level~1 and the short-navigation task in Level~2.
They change visual evidence at the same task locations.
These variations provide the tasks for RQ3, which examines adaptation while presupposing the earlier capabilities.

\subsection{Human Annotation}
\label{app:human-annotation}

We manually construct and verify each evaluation instance in \UrbanGroundName{}.
For each instance, annotators execute the task from its specified initial state using the same tools and constraints as the evaluated agents, and confirm that it can be completed within the 150-step limit.
Every instance then undergoes a second round of verification, and identified problems are corrected before inclusion.

\begin{figure*}[t!]
    \centering
    \begin{minipage}[t!]{0.59\textwidth}
        \vspace{0pt}
        \centering
        \includegraphics[width=\linewidth]{Figures/task_distribution}
        \vspace{-1.5mm}
        \captionof{figure}{Spatial distribution of experimental tasks across Hong Kong.}
        \label{fig:task-distribution}
    \end{minipage}
    \hfill
    \begin{minipage}[t!]{0.40\textwidth}
        \vspace{0pt}
        \centering
        \includegraphics[width=\linewidth]{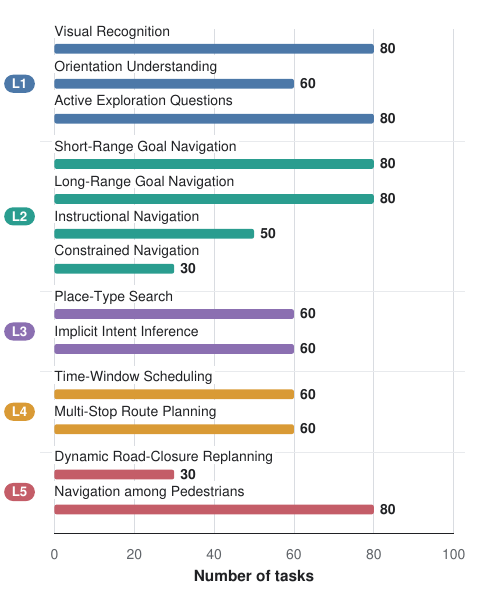}
        \vspace{-1.5mm}
        \captionof{figure}{Number of task instances at each level of the spatial agency.}
        \label{fig:task-composition}
    \end{minipage}
    \vspace{-2mm}
\end{figure*}

Our annotation covers all major regions across Hong Kong, including areas with different street patterns, terrain, and vertical pedestrian connections.
Figure~\ref{fig:task-distribution} shows the geographic distribution of the annotated tasks, while Figure~\ref{fig:task-composition} summarizes the number of instances at each level of the spatial agency ladder.
Complete task definitions and construction details are provided in Appendix~\ref{sec: Experimental Task Design}.

\input{sections/experiments}

\bibliography{iclr2027_conference}
\bibliographystyle{iclr2027_conference}

\clearpage
\appendix
\section{Comparison with Urban Environments}
\label{app:urban_environment_comparison}

Table~\ref{tab:urban_environment_comparison} compares the urban environments and navigation benchmarks. 
Real-city geography denotes a layout or navigation graph tied to an existing city, including layouts derived from OpenStreetMap. 
Continuous motion refers to movement through a spatially continuous scene, including motion executed through discrete action commands.
Collision response requires scene geometry to constrain executed movement. 
Pedestrian Network refers to a georeferenced 3D graph that represents real-world pedestrian paths and their connectivity across elevation levels. 
Locomotion primitives control denotes evaluation in which the model selects locomotion primitives without a separate navigation planner.

\begin{table}[t!]
\centering
\caption{Comparison of urban environments and navigation benchmarks in Section~\ref{sec: Related Work}. 
\UGYes\ documented support, 
\UGNo\ absent from the described setting, 
}
\label{tab:urban_environment_comparison}

\small
\setlength{\tabcolsep}{2.5pt}
\renewcommand{\arraystretch}{1.25}

\begin{tabular*}{\linewidth}{@{\extracolsep{\fill}}lccccc@{}}
\toprule
\textbf{Environment / benchmark}
& \makecell{\textbf{Real-city}\\
            \textbf{geography}}
& \makecell{\textbf{Continuous}\\
            \textbf{motion}}
& \makecell{\textbf{Collision}\\
            \textbf{response}}
& \makecell{\textbf{3D Pedestrian}\\
            \textbf{network}}
& \makecell{\textbf{Locomotion}\\
            \textbf{primitives}} \\
\midrule

V-IRL~\citep{V-IRL}
& \UGYes & \UGNo & \UGNo & \UGNo & \UGNo \\

CitySeeker~\citep{CitySeeker}
& \UGYes & \UGNo & \UGNo & \UGNo & \UGNo \\

CityNav~\citep{Exploring_Emergent_Navigation}
& \UGYes & \UGNo & \UGNo & \UGNo & \UGNo \\

\addlinespace[3pt]

UrbanWorld~\citep{UrbanWorld}
& \UGYes & \UGYes & \UGNo & \UGNo & \UGNo \\

MetaUrban~\citep{MetaUrban}
& \UGNo & \UGYes & \UGYes & \UGNo & \UGNo \\

SimWorld~\citep{SimWorld}
& \UGNo & \UGYes & \UGYes & \UGNo & \UGNo \\

SimWorld-Robotics~\citep{SimWorld-Robotics}
& \UGNo & \UGYes & \UGYes & \UGNo & \UGYes \\

Virtual Community~\citep{Virtual_Community}
& \UGYes & \UGYes & \UGYes & \UGNo & \UGNo \\

\midrule

\textbf{\UrbanGroundName{} (ours)}
& \UGYes & \UGYes & \UGYes & \UGYes & \UGYes \\

\bottomrule
\end{tabular*}

\end{table}

\section{Experimental Task Design}
\label{sec: Experimental Task Design}

Section~\ref{sec:spatial-agency-ladder} introduces the five-level evaluation ladder and its relation to the research questions.
This appendix records how individual tasks were constructed and gives the complete definitions behind each level.

Each instance is manually constructed and verified in \UrbanGroundName{}.
Annotators execute every task and confirm that it can be completed from the specified initial state with the provided tools and constraints.
Every instance then undergoes a second round of verification, and problems are corrected before inclusion.
Annotators also vary the tasks and distribute instances across different regions of Hong Kong.
The sequence follows the point at which local judgments must be attached to a wider spatial frame, from the immediate surroundings to long-range objectives, uncertain destinations, temporal constraints, and environmental changes.

\cellcolor{LevelOneLight}\Lone{\subsection{Level 1: Local Environment Understanding}}

Level~1 evaluates whether an agent can establish a local spatial understanding before undertaking long-range movement. 
The tasks examine scene recognition, orientation, and information gathering within the immediate surroundings.

\paragraph{Visual recognition (VR).}
The agent identifies task-relevant urban content from the visual observations available at the initial location. 
It must associate visible scene evidence with the semantic category queried by the task and select the corresponding answer. 
Locomotion is not required, although the agent may change its viewing direction to inspect the surrounding scene.

\paragraph{Orientation understanding (OU).}
The agent recovers directional relations from the local visual configuration. 
It must establish a consistent spatial frame from the current viewpoint and use that frame to determine the orientation or relative position requested by the task. 
Correct completion requires preserving the correspondence between the agent's heading, the observed scene layout, and the referenced urban locations.

\paragraph{Active exploration questions (AEQ).}
The agent answers a local spatial question whose evidence is not fully available at the initial pose. 
It must determine which additional observations are needed, move to informative viewpoints, and integrate the collected evidence into a single answer. 
The exploration remains confined to the nearby environment and is designed so that a 60-second short interaction is sufficient when the agent chooses an effective observation strategy.

\cellcolor{LevelTwoLight}\Ltwo{\subsection{Level 2: Navigation under Explicit Instructions}}

Level~2 evaluates whether the agent can execute a navigation objective whose destination or route information is explicitly provided. 
The subtasks vary in travel distance, route representation, and the presence of constraints, while keeping the intended goal known to the agent.

\paragraph{Short-range goal navigation (SGN).}
The agent moves from the initial pose to a visible destination using first-person control.  The map tool is not needed, and successful execution depends on maintaining local visual alignment throughout the movement.

\paragraph{Long-range goal navigation (LGN).}
The agent reaches a destination that is not visible from the initial pose and lies beyond the immediately observed surroundings. 
It must use the available map information to establish a global route and repeatedly align that route with its current first-person observations. 
Completion requires sustained localization and route following across an extended trajectory. 
The journey often requires exploring several streets.

\paragraph{Instructional navigation (IN).}
The agent receives a sequential route description whose individual directives become grounded only as execution progresses. 
The final target location is not provided as a directly queryable coordinate. The agent must maintain its position within the instruction sequence and execute the corresponding transition before advancing to the next directive. 
Success requires the complete route description to remain synchronized with the observed environment.

\paragraph{Constrained navigation (CN).}
The agent reaches an explicit destination while respecting restrictions on which parts of the pedestrian environment may be used. 
It must incorporate these restrictions into route selection and avoid trajectories that violate the specified constraints.

\cellcolor{LevelThreeLight}\Lthree{\subsection{Level 3: Exploration under Implicit Instructions}}

Level~3 removes direct target specification. 
The agent must infer an appropriate destination from the instruction and then complete the corresponding navigation.

\paragraph{Place-type search (PTS).}
The agent is given a destination category without a designated endpoint. 
It must locate a valid instance within the surrounding city and navigate to it. 
The task therefore couples semantic search with spatial exploration.

\paragraph{Implicit intent inference (III).}
The instruction describes an intended outcome without naming the place that can realize it. 
The agent first resolves the instruction into a concrete urban destination, then completes the corresponding navigation. 

\cellcolor{LevelFourLight}\Lfour{\subsection{Level 4: Multi-Task Planning}}

Level~4 evaluates urban exploration when a single episode contains several destinations. 
The central difficulty shifts from reaching one endpoint to maintaining a coherent plan over a sequence of visits.

\paragraph{Time-window scheduling (TWS).}
The agent receives multiple destinations governed by temporal constraints. 
It must derive a feasible schedule from the available time intervals and execute the resulting sequence under the \UrbanGroundName{} clock. 
A trajectory succeeds only when each visit occurs within its assigned window, so route selection is evaluated together with temporal feasibility.

\paragraph{Multi-stop route planning (MSP).}
The agent is required to visit a set of destinations whose order is left unspecified. 
It must determine an efficient visiting sequence from their geographic arrangement and carry that sequence through continuous navigation.

\cellcolor{LevelFiveLight}\Lfive{\subsection{Level 5: Dynamic Environment Interaction}}

Level~5 evaluates whether a navigation policy remains effective after the environment departs from its initial state. 
The tasks are derived from validated navigation instances and introduce changes during execution without altering the original objective.

\paragraph{Dynamic road-closure replanning (DCR).}
A pedestrian segment becomes unavailable after navigation has begun.
A one-time system notice identifies the closure, which remains marked on the map.
The previously selected route may therefore cease to be executable before the destination is reached.
The agent must revise its route and continue through a valid alternative.

\paragraph{Navigation among pedestrians (NP).}
Moving pedestrians are introduced into validated long-range navigation instances from the open-source Microsoft Rocketbox avatar library~\citep{Rocketbox}. 
The agent must preserve progress toward the destination while adjusting its motion to avoid contact with pedestrians.

\input{sections/full_results_appendix}

\input{appendix_prompts}

\end{document}

%% file: math_commands.tex
\usepackage{amsmath,amsfonts,bm}

\def\eqref#1{equation~\ref{#1}}

\def\1{\bm{1}}

\DeclareMathAlphabet{\mathsfit}{\encodingdefault}{\sfdefault}{m}{sl}
\SetMathAlphabet{\mathsfit}{bold}{\encodingdefault}{\sfdefault}{bx}{n}



%% file: paper_revision.tex
\usepackage{threeparttable}

\colorlet{RQOne}{LevelOne}
\colorlet{RQOneLight}{LevelOneLight}
\colorlet{RQTwo}{LevelTwo}
\colorlet{RQTwoLight}{LevelTwoLight}
\colorlet{RQThree}{LevelFive}
\colorlet{RQThreeLight}{LevelFiveLight}

\newtcolorbox{rqbox}[3]{
  enhanced,
  breakable,
  colback=#1Light!58!white,
  colframe=#1,
  boxrule=0.55pt,
  arc=1.8mm,
  boxsep=0pt,
  left=2.2mm,
  right=2.2mm,
  top=1.6mm,
  bottom=1.5mm,
  before skip=0.55em,
  after skip=0.55em,
  title={\sffamily\bfseries\color{#1}#2\quad\color{black!86}#3},
  colbacktitle=#1Light,
  coltitle=#1,
  fonttitle=\normalsize,
  attach boxed title to top left={xshift=1.5mm,yshift*=-1.0mm},
  boxed title style={boxrule=0pt,arc=1.2mm,left=1.1mm,right=1.1mm,top=0.6mm,bottom=0.6mm}
}

\newcommand{\rqsection}[2]{%
  \subsection{\texorpdfstring{\textcolor{#1}{#2}}{#2}}%
}

\newif\ifurbanrevised
\urbanrevisedtrue

\newcommand{\UrbanGroundHeroFigure}{%
  \ifurbanrevised
    \begin{center}
      \includegraphics[width=0.975\textwidth]{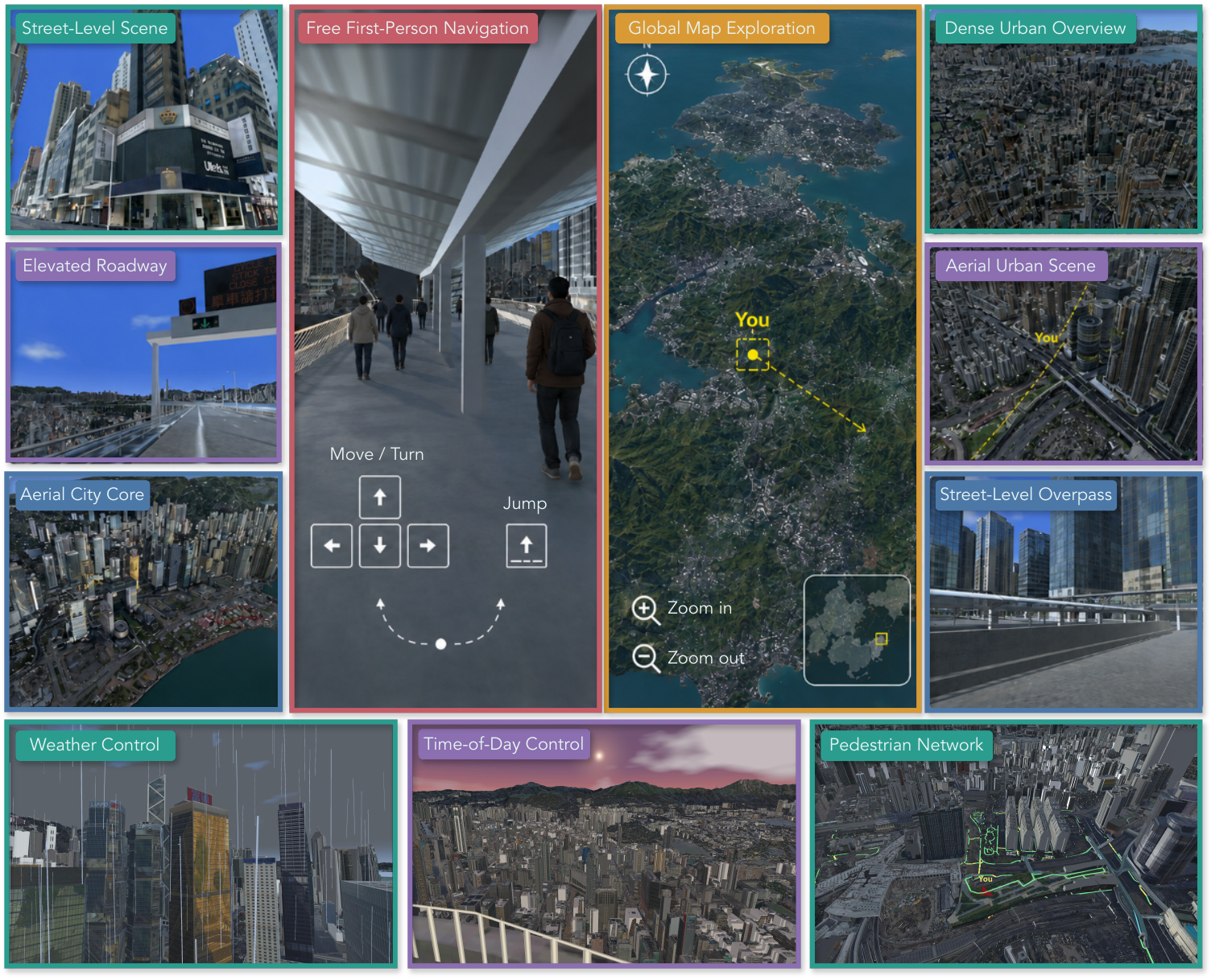}
      \captionsetup{type=figure}
      \captionof{figure}{\UrbanGroundName{} is a real-scale urban sandbox built from territory-wide 3D geospatial data. It supports direct first-person play and programmatic control by MLLM agents with movement constrained by physical collision geometry. 
      It also provides a shared evaluation interface for studying how MLLM agents use local observations to act across the city.}
      \label{fig:UrbanGround}
    \end{center}
    \vspace{-0.35em}
  \fi
}

%% file: sections/abstract.tex
Multimodal large language models (MLLMs) can interpret a street view, but reliable urban action depends on whether such local evidence remains useful after the agent starts to move.
In this paper, we investigate \emph{how far current MLLM agents can turn local urban perception into reliable action in a real-scale city.}
We propose \UrbanGroundName{}, an urban sandbox built from Hong Kong's territory-wide 3D geospatial data. 
It combines the city's geographic structure with continuous, collision-constrained control through a shared evaluation interface. 
Agents use first-person observations and an interactive map to select actions across tasks ranging from local question answering to long-horizon navigation.
Our analysis follows the growth of the spatial problem through three research questions.
We first test whether an agent can gather and interpret local visual evidence to answer spatial questions. 
Then we ask whether these abilities support navigation as destinations become farther away and less explicit. 
Finally, we examine whether the resulting behavior survives changes in route availability and pedestrian motion. 
MLLM agents usually show useful atomic abilities in visual recognition and short-range spatial reasoning, while orientation and pedestrian-aware movement remain unreliable.
Their central failure emerges over extended exploration, where local abilities do not compose into sustained goal-directed behavior and errors accumulate without effective correction.
We hope \UrbanGroundName{} will support broader study of how far MLLM agents can explore reliably in open-ended urban environments. 

%% file: sections/introduction.tex
\section{Introduction}

\begin{tcolorbox}[
  colback=LevelOne!8,
  colframe=LevelOne!90,
  boxrule=0.55pt,
  arc=2.2mm,
  boxsep=0pt,
  left=2.5mm,
  right=2.5mm,
  top=2.0mm,
  bottom=1.7mm,
  before skip=0.6em,
  after skip=0.9em
]
\raggedright\itshape
``Nothing is experienced by itself, but always in relation to its surroundings, the sequences of events leading up to it, the memory of past experiences.''%
\par\smallskip
\raggedleft
Kevin Lynch, \textit{The Image of the City} (1960)
\end{tcolorbox}

Recent multimodal large language model (MLLM) agents have shown strong capabilities in recognizing urban objects from individual observations~\citep{MLLM_Agent_Survey,UrbanLLaVA}.
However, agents receive these observations through a continuing physical process in which each movement produces the next first-person view~\citep{NaVILA, Embodied_Model}.
Once the agent turns a corner, a landmark that supported its last decision may disappear even though its location still matters.
Earlier evidence then has to be reconciled with the new view so that the agent retains a usable estimate of its pose.
Strong recognition at one viewpoint therefore does not show whether spatial understanding will remain coherent through continued action~\citep{CitySeeker, VLN-MME}.

Recent work has approached this problem by placing agents in progressively longer interactions across larger spaces.
Game sandboxes support extended trajectories through changing worlds~\citep{VideoGameBench,lmgame-Bench,luo2026aigamesfoundationmodel}.
They reveal how decisions unfold over time, yet success remains shaped by game-specific mechanics~\citep{MineExplorer}.
Moving into physically grounded settings makes each action produce a new observation from the updated pose under scene geometry~\citep{OpenEQA,NavBench,CapNav}.
Their bounded spaces rarely require a spatial account to remain useful beyond a local area.
Urban studies extend the scale through aerial imagery panoramas~\citep{VELMA,Exploring_Emergent_Navigation,CitySeeker}.
These observations preserve real geographic context, but the agent still moves between sampled viewpoints without continuous physical contact.
Interactive urban simulators restore the action-observation loop, although their scenes do not preserve the full complex geographic structure of an existing metropolis~\citep{UrbanWorld,MetaUrban,SimWorld,Virtual_Community}.

We therefore ask: \emph{how far can MLLM agents turn local urban perception into reliable action in a real-scale physical city?}
At first, the agent only needs to establish where relevant evidence lies around its current pose.
It must then carry this estimate beyond the visible scene and use it to control physical movement.
Once the city changes, the agent has to revise the estimate without losing the goal. 
A useful evaluation should expose each point in this transition under one geographic frame.

In this paper, we present \UrbanGroundName{}, a real-scale environment that turns Hong Kong's complicated geographic structure into an interactive world for MLLM agents.
It is built from the 3D Visualisation Map and 3D Pedestrian Network released by the Lands Department of the Government of the Hong Kong Special Administrative Region\footnote{Official dataset pages: \href{https://data.gov.hk/en-data/dataset/hk-landsd-openmap-3d-visualisation-map-tile-based-models}{3D Visualisation Map} and \href{https://data.gov.hk/en-data/dataset/hk-landsd-openmap-3d-pedestrian-network}{3D Pedestrian Network}.}.
Dense development sits on steep terrain, while the pedestrian network repeatedly shifts between street level and elevated passages. 
Moving toward the goal can therefore lead to an inaccessible approach. 
Agents must combine local observations with map context to infer which pedestrian connections provide access 
(Section~\ref{sec: Implementation of UrbanGround}).

We study this problem through three research questions that track the growth of spatial demands:

\begin{rqbox}{RQOne}{RQ1}{Can MLLM agents establish a usable local spatial grounding?}
We begin with active spatial question answering, where the agent can move briefly to gather visual evidence. 
Current MLLMs perform well on these local tasks, suggesting that they already possess useful atomic skills for urban spatial understanding. 
However, answer accuracy does not guarantee physically valid behavior. 
Agents sometimes leave the pedestrian network even after reaching the answer (Section~\ref{sec: RQOne}).
\end{rqbox}

\begin{rqbox}{RQTwo}{RQ2}{How does local grounding scale into goal-directed navigation?}
We then extend the evaluation from local exploration to navigation beyond the visible scene. 
Agents can often reach a nearby destination that is visible from the starting pose, but performance collapses when the route spans only a few city blocks. 
Agents may identify the correct global direction and still fail to find a feasible path through the local geometry. 
Once blocked, they often repeat the same action without recovering, exposing a sharp gap between spatial recognition and sustained control (Section~\ref{sec: RQTwo}).
\end{rqbox}

\begin{rqbox}{RQThree}{RQ3}{How do changes in the city affect grounded perception and action?}
We finally keep the task objective fixed while changing the city around the agent.
Reduced visibility mainly weakens local question answering, while its effect on short-range navigation is less consistent across models.
When a route becomes invalid during execution, agents often continue to produce locally compliant movements on the pedestrian network even after their spatial plan has become obsolete. 
Plausible local behavior does not ensure effective adaptation to changes in route availability 
(Section~\ref{sec: RQThree}).
\end{rqbox}

Overall, our results show that sustained goal-directed behavior cannot be inferred from isolated perception or success over a short route. 
Current MLLMs possess useful local spatial skills for interpreting nearby scenes, but these skills lose reliability as interaction extends through the city. 
\UrbanGroundName{} makes this gap observable in a real geographic setting, where every action changes the evidence available to the agent. 
We hope \UrbanGroundName{} provides a foundation for studying how local perception can support sustained, physically grounded behavior at city scale.

%% file: sections/experiments.tex
\section{Experiments}
\label{sec: Experiments}

\newcommand{\casepanel}[5]{%
\begin{minipage}[t]{0.238\textwidth}
\centering
{\setlength{\fboxsep}{0pt}\setlength{\fboxrule}{0.45pt}%
\fcolorbox{#1}{white}{%
\includegraphics[width=\dimexpr\linewidth-2\fboxrule\relax]{#2}}}
\par\vspace{0.22em}
{\raggedright
(#3) Step #4, #5\par}
\end{minipage}%
}

\subsection{Experimental Setup}

\paragraph{Models.}
We evaluate contemporary MLLMs from several model families.
The GPT family includes GPT-5.6-Luna, GPT-5.5, GPT-5.4, and GPT-5.2~\citep{GPT-5.6,GPT-5.5,GPT-5.4,GPT-5.2}.
The Claude family includes Claude-Opus-5 and Claude-Opus-4.6~\citep{Claude-Opus-5,Claude-Opus-4.6}.
The Gemini family includes Gemini-3.8-Flash, Gemini-3.6-Flash and Gemini-3.1-Pro~\citep{Gemini-3.8-Flash,Gemini-3.1-Pro}.
Additional evaluated models are Doubao-Seed-2.0-Pro~\citep{Seed2.0}, GLM-5V-Turbo~\citep{GLM-5V-Turbo}, and Kimi-K3~\citep{Kimi-K3}.
The main tables show representative models, while Appendix~\ref{app:full-results} provides the complete matrices.

\paragraph{Agent protocol.}
At each step, the agent receives the current first-person RGB observation, the task instruction, and its text interaction history.
It selects one structured action for execution in the simulator from the physical or map action space in Section~\ref{sec: Agent Layer}.
An episode ends when the agent submits an answer, signals completion, or reaches 150 interaction steps.
The prompts are fixed within each task type and are reported in Appendix~\ref{app:prompts}.

\paragraph{Tasks and metrics.}
The study contains 810 manually verified base instances distributed across Hong Kong and organized by the evaluation ladder in Section~\ref{sec:spatial-agency-ladder}.
Every instance was completed by human testers under the same 150-step limit used for MLLM agents, including some tool-calling time (Section~\ref{app:human-annotation}).
Since each movement action lasts at most two seconds, this budget permits up to 300 seconds of commanded motion, so model failures cannot be attributed to tasks that are infeasible within the evaluation horizon.
Question-answering episodes are scored by agreement with the annotated answer.
Single-endpoint navigation with a labeled target succeeds when the final position is within 15 meters of the destination and any evaluator-enforced task constraint is satisfied.
Completion signals for place search and multi-destination tasks are defined with their results.
We report pedestrian-network adherence as the fraction of total action time whose post-action state lies on the registered pedestrian network.
Dynamic tasks add event-specific records, including closure violations, pedestrian contact, and rain exposure.

\rqsection{RQOne}{RQ1: Can MLLM agents establish a usable local spatial grounding?}
\label{sec: RQOne}

We begin by testing whether an agent can establish a spatial state before it must use that state over a route.
Visual recognition asks the agent to identify task-relevant content in its initial surroundings.
Orientation tests whether visible places can be located in a stable heading-relative frame.
Active exploration requires the agent to seek evidence that is missing from the initial pose before answering.
Table~\ref{tab:rq1-grounding} reports answer accuracy and pedestrian-network adherence.
\begin{table}[t!]
\centering
\caption{Answer accuracy and pedestrian-network adherence compare recognition, orientation, and active exploration question answering across the evaluated MLLM agents. Overall is the instance-count-weighted average across the three task types.}
\label{tab:rq1-grounding}
\begin{threeparttable}
\setlength{\tabcolsep}{2.0pt}
\renewcommand{\arraystretch}{1.12}
\begin{tabularx}{\textwidth}{@{}l*{8}{>{\centering\arraybackslash}X}@{}}
\toprule
\multirow{2}{*}{\textbf{Model}} &
\multicolumn{2}{c}{\makebox[0pt][c]{\textbf{Visual Recognition}}} &
\multicolumn{2}{c}{\makebox[0pt][c]{\textbf{Orientation}}} &
\multicolumn{2}{c}{\makebox[0pt][c]{\textbf{Active Exploration}}} &
\multicolumn{2}{c}{\makebox[0pt][c]{\textbf{Overall}}} \\
\cmidrule(lr){2-3}\cmidrule(lr){4-5}\cmidrule(lr){6-7}\cmidrule(lr){8-9}
 & Acc. & PNA & Acc. & PNA & Acc. & PNA & Acc. & PNA \\
\midrule
GPT-5.6-Luna & 78.8 & 64.7 & 48.3 & 71.6 & 58.8 & 92.2 & 63.2 & 76.6 \\
GPT-5.5 & 82.5 & 63.8 & 40.0 & 74.8 & 62.5 & 91.4 & 63.6 & 76.8 \\
GPT-5.4 & 75.0 & 66.2 & 31.7 & \underline{75.0} & 57.5 & 91.1 & 56.8 & 77.7 \\
GPT-5.2 & 77.5 & 68.0 & 33.3 & 74.7 & 48.8 & 90.9 & 55.0 & 78.2 \\
Claude-Opus-5 & 91.3 & \underline{69.4} & \textbf{58.3} & \textbf{77.9} & \textbf{82.5} & 92.8 & \textbf{79.1} & \textbf{80.2} \\
Claude-Opus-4.6  & 85.0 & \textbf{70.0} & 46.7 & \underline{75.0} & 73.8 & 92.5 & 70.5 & \underline{79.5} \\
Gemini-3.8-Flash & \underline{92.5} & 63.1 & \textbf{58.3} & 69.8 & \underline{80.0} & \textbf{94.1} & \underline{78.6} & 76.2 \\
Gemini-3.6-Flash & \textbf{93.8} & 68.9 & 56.7 & 74.3 & 77.5 & 92.5 & 77.7 & 79.0 \\
Gemini-3.1-Pro & 82.5 & 65.2 & 23.3 & 59.1 & 46.3 & \underline{93.0} & 53.2 & 73.6 \\
Doubao-Seed-2.0-Pro & 85.0 & 64.0 & 36.7 & 73.2 & 63.8 & 91.1 & 64.1 & 76.4 \\
GLM-5V-Turbo & 80.0 & 67.5 & 36.7 & 71.2 & 52.5 & 90.6 & 58.2 & 76.9 \\
Kimi-K3 & \underline{92.5} & 66.2 & 55.0 & 67.4 & 76.3 & 91.2 & 76.4 & 75.6 \\
\bottomrule
\end{tabularx}
\end{threeparttable}
\end{table}

\textbf{Current MLLM agents already show reliable performance on short-range atomic spatial reasoning tasks.}
Most models perform strongly on visual recognition when both the instruction and the relevant evidence are explicit.
They can accurately understand the nearby visual environment and complete the atomic grounding task.
When active exploration requires actions to uncover missing evidence, performance varies substantially across models. 
Accuracy ranges from 46.3\% to 82.5\%, showing useful but uneven evidence-gathering abilities. 
Some agents handle exploratory sequences well, whereas others struggle despite clear questions with explicit answer choices.

\textbf{However, compared with reasoning about nearby spatial objects, MLLM agents remain considerably less reliable when answering questions that require directional awareness.}
Performance falls markedly across models, and several systems approach the random-guessing baseline of 25\% answer accuracy for a four-option question.
The contrast with visual recognition shows that identifying visible landmarks does not ensure a stable representation of their directions relative to the current heading across successive observations.
Directional grounding is therefore a shared atomic weakness that appears before long-horizon navigation is required.

Interestingly, the gap between earlier and more recent generations of MLLMs emerges most clearly in orientation understanding and active exploration. 
GPT-5.2 and Gemini-3.1-Pro remain close to newer models in visual recognition, yet fall markedly behind on the other two tasks. 
This suggests that recent advances have primarily improved the ability to preserve directional spatial evidence, bringing MLLMs closer to turning local perception into spatial agency.

Pedestrian-network adherence further exposes a surprising gap between accurate answers and road-compliant movement.
We measure the fraction of action time that the agents remain on registered pedestrian routes.
Adherence falls far short of complete compliance during several local tasks.
The agents often favor a shorter path to the necessary evidence and sacrifice road compliance in order to answer the user more quickly.
It suggests that real-world movement constraints are not maintained as a persistent part of spatial reasoning when they compete with rapid task completion.

\begin{figure}[t!]
\centering
\casepanel{RQOne}{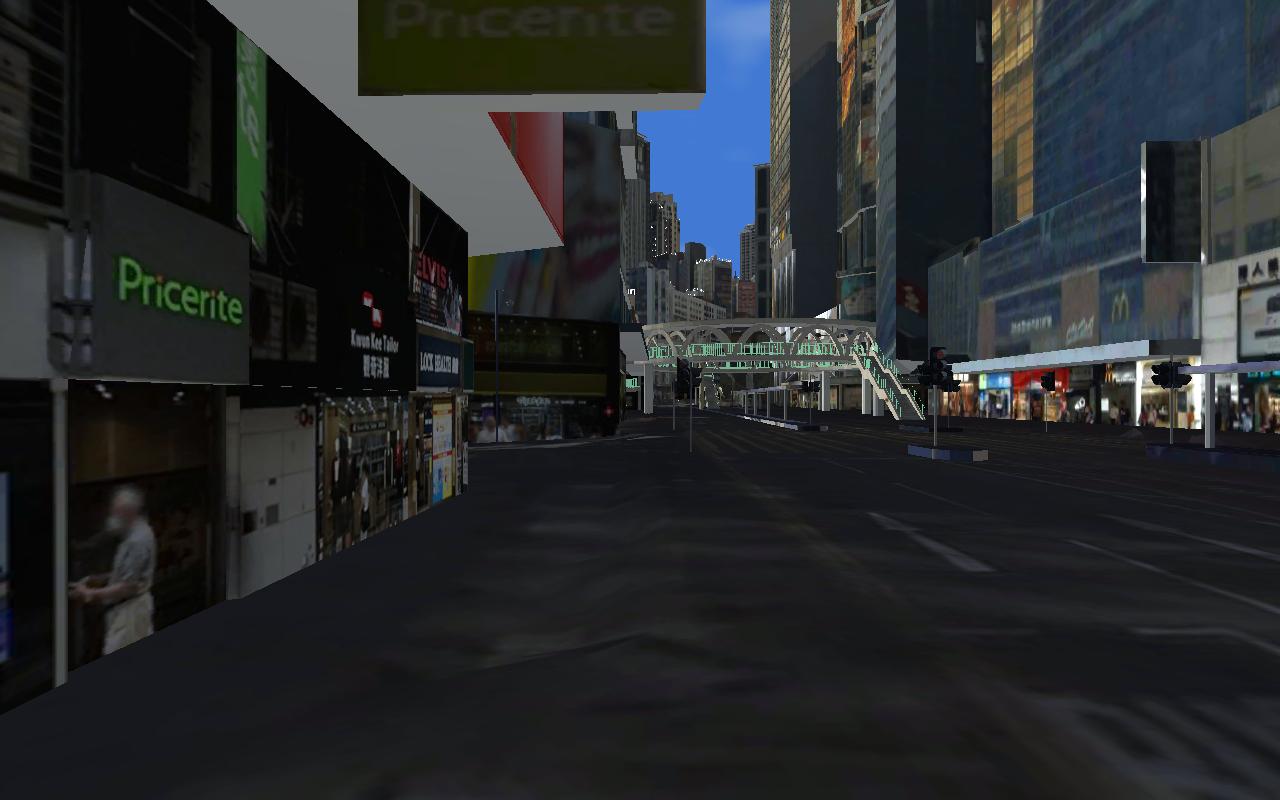}{a}{1}{look right $60^\circ$, pitch 0}\hfill
\casepanel{RQOne}{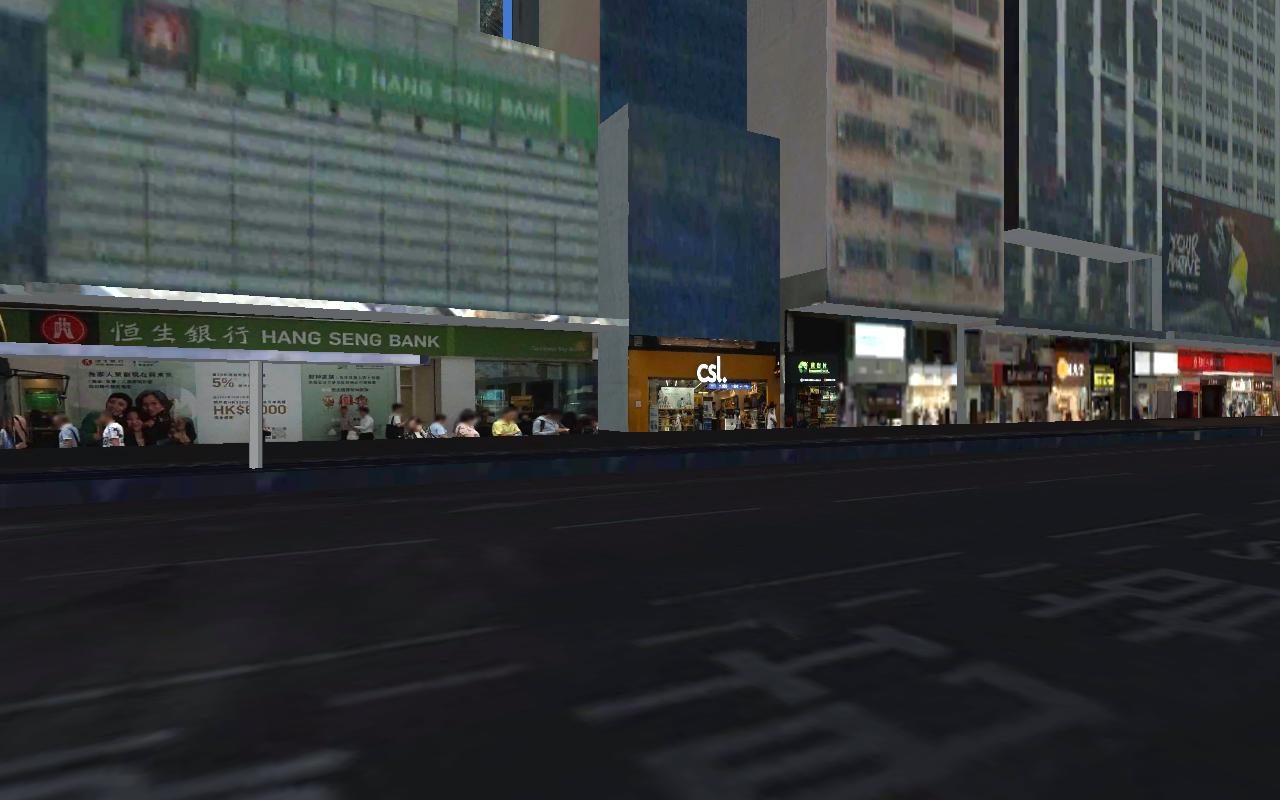}{b}{3}{look right $45^\circ$, pitch 0}\hfill
\casepanel{RQOne}{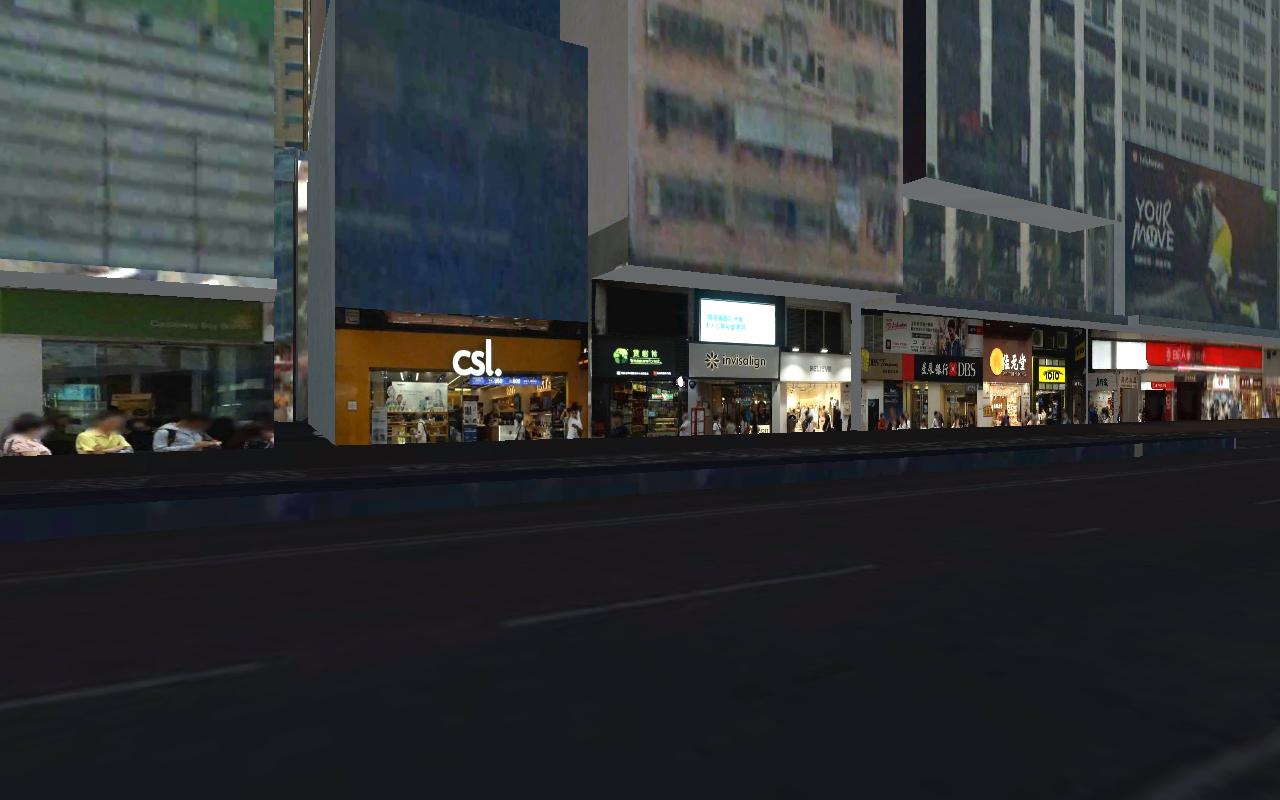}{c}{6}{look right $25^\circ$, pitch 0}\hfill
\casepanel{RQOne}{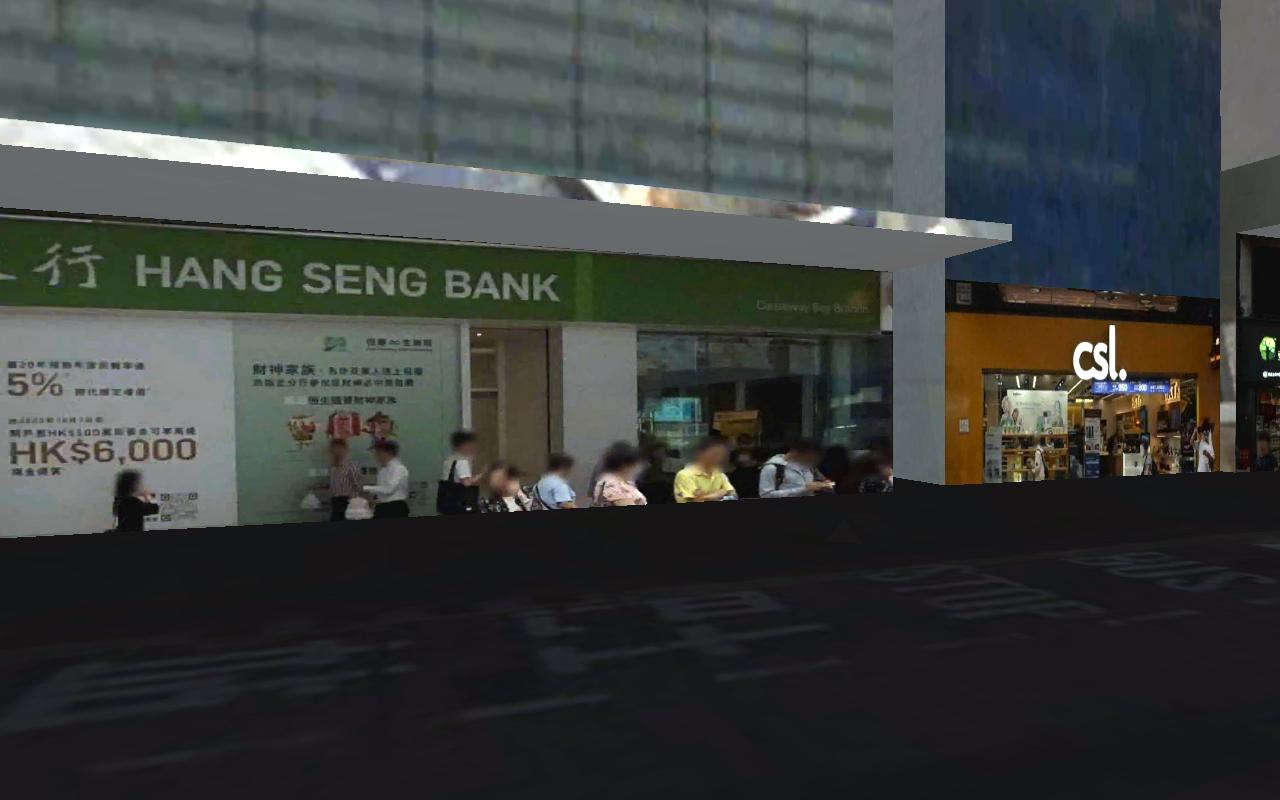}{d}{9}{sprint
forward 2.0 s}
\par\smallskip
\casepanel{RQOne}{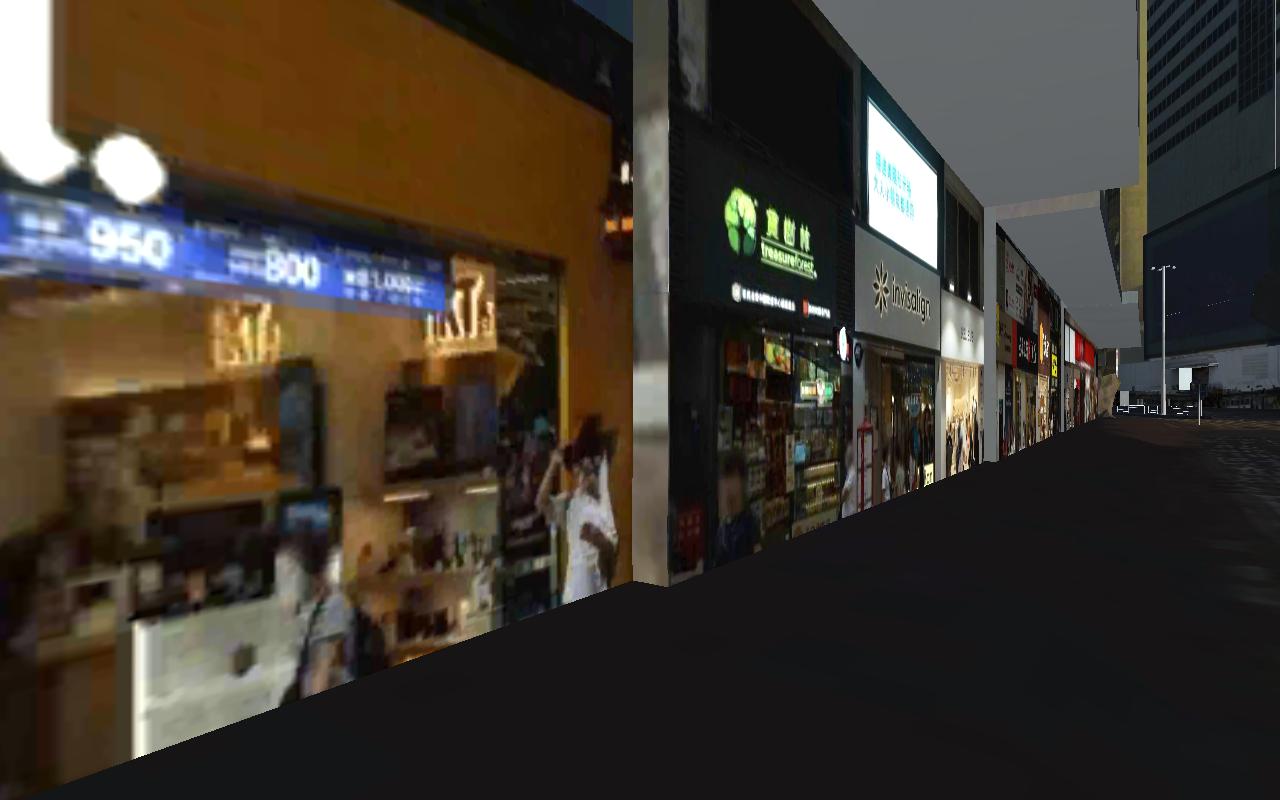}{e}{17}{look left $25^\circ$, pitch 0}\hfill
\casepanel{RQOne}{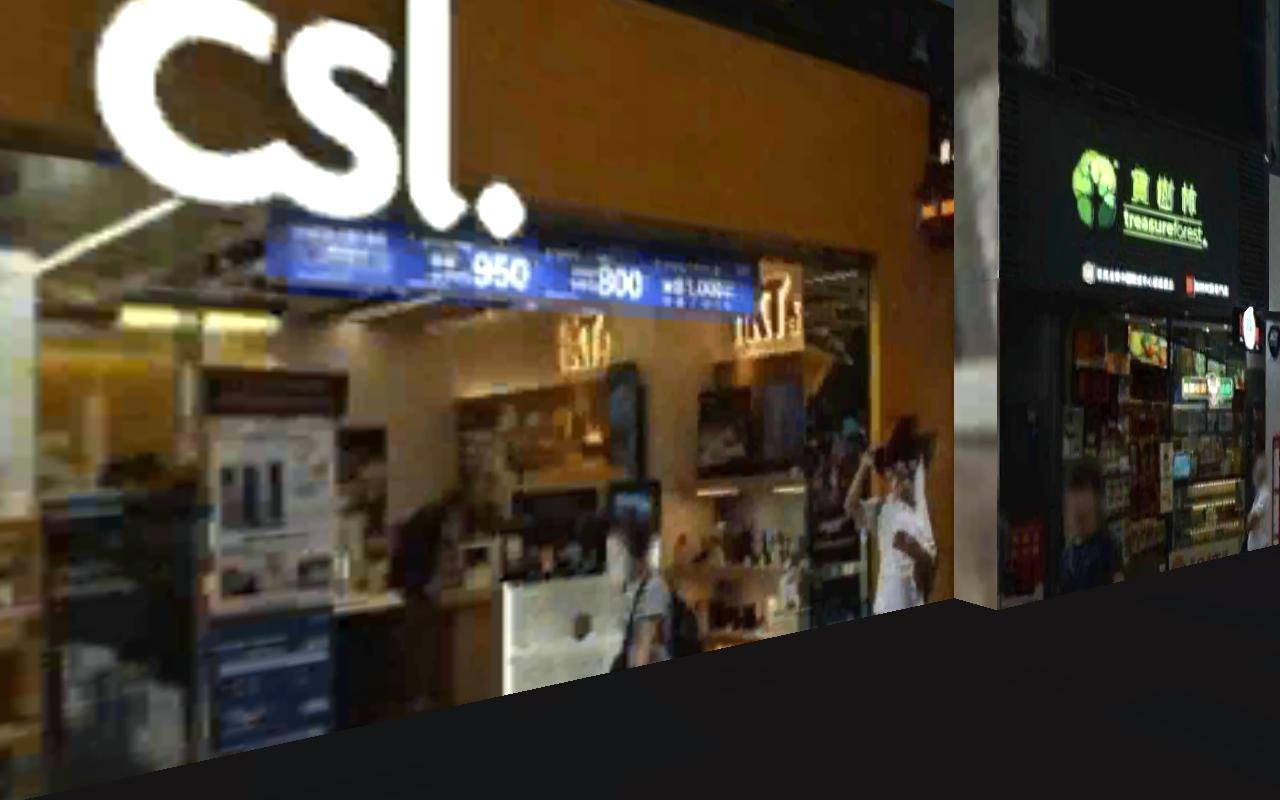}{f}{18}{look right $35^\circ$, pitch 0}\hfill
\casepanel{RQOne}{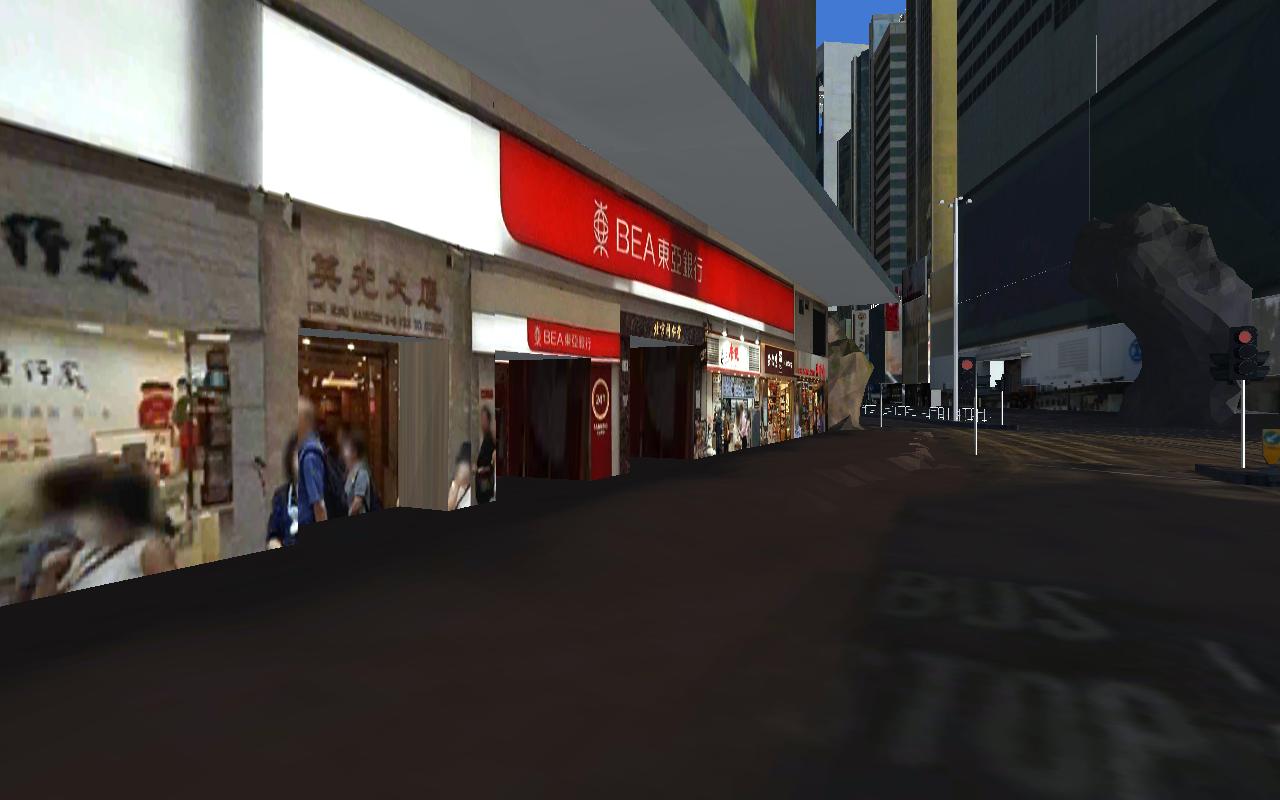}{g}{25}{look left $15^\circ$, pitch 0}\hfill
\casepanel{RQOne}{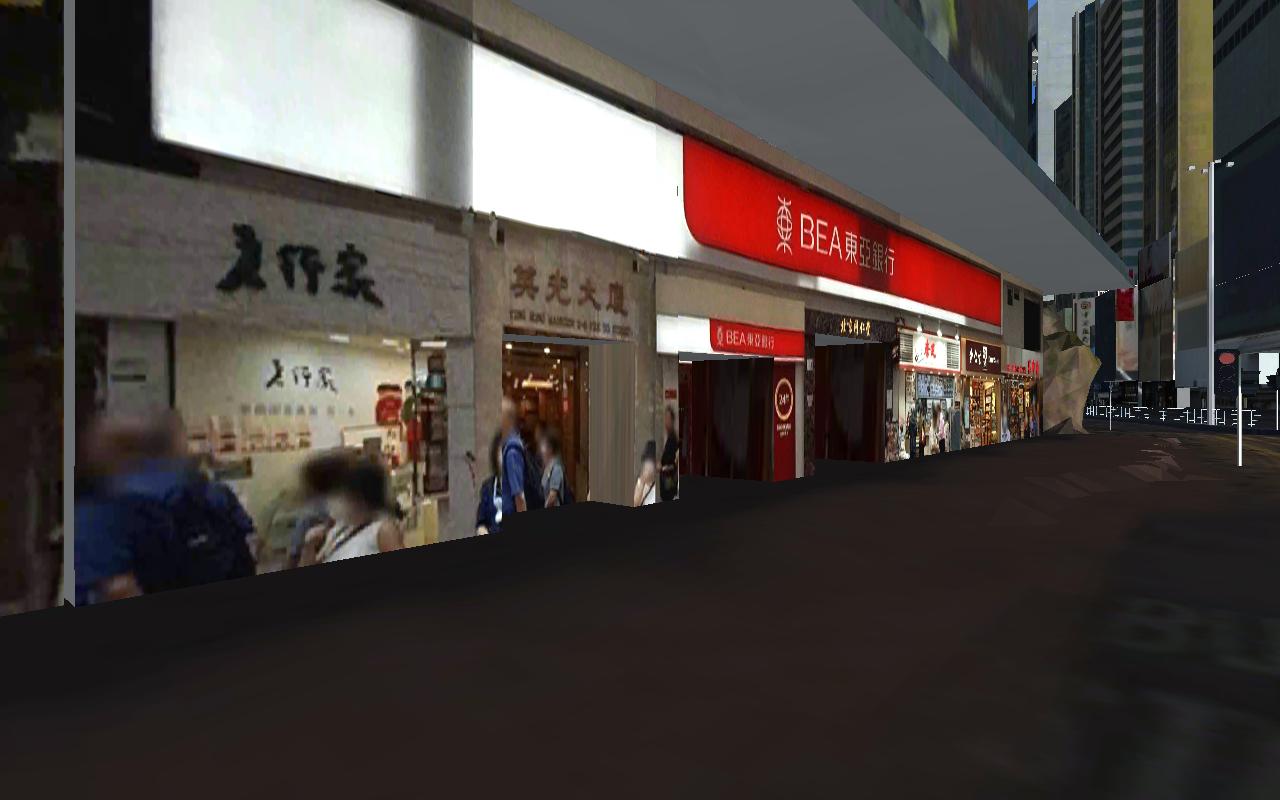}{h}{26}{terminate}
\caption{Example of active exploration by GPT-5.5. The agent is asked to answer which bank is next to Beijing Tong Ren Tang.}
\label{fig:rq1-bank-case}
\end{figure}

Figure~\ref{fig:rq1-bank-case} provides an example of GPT-5.5 performing active exploration after being asked to identify the bank next to Beijing Tong Ren Tang.
The agent successfully locates the pharmacy and the neighboring bank, approaches the storefront, and answers Bank of East Asia.
However, to finish quickly, the agent crosses the road directly and leaves the registered pedestrian connection.
The episode shows that the model can ground the visible target well enough to answer while its understanding of pedestrian affordances and real road constraints remains limited.

\rqsection{RQTwo}{RQ2: How does local grounding scale into goal-directed navigation?}
\label{sec: RQTwo}

We next test whether local grounding survives when the agent must maintain a goal over a route.
Table~\ref{tab:rq2-success} covers eight navigation tasks that progressively increase the spatial state required during execution.
Short and long navigation provide a named endpoint at different route scales.
Instructional navigation grounds a sequence of verbal directives, while constrained navigation restricts the usable path.
Place search and intent inference require the destination to be inferred.
Time-window and multi-stop tasks require several visits to remain active within one plan.
\begin{table}[t!]
\centering
\caption{Navigation success across tasks grouped by whether the destination is provided, inferred, or multiple, where SN, LN, IN, CN, PS, II, TW, and MS denote short navigation, long navigation, instructional navigation, constrained navigation, place search, intent inference, time-window navigation, and multi-stop navigation, respectively, and Overall is the instance-count-weighted average across all eight subtasks.}
\label{tab:rq2-success}
\setlength{\tabcolsep}{1.5pt}
\renewcommand{\arraystretch}{1.10}
\begin{tabular*}{\textwidth}{@{\extracolsep{\fill}}l*{9}{>{\centering\arraybackslash}p{3.1em}}@{}}
\toprule
\multirow{2}{*}{\textbf{Model}} &
\multicolumn{4}{c}{\textbf{Provided}} &
\multicolumn{2}{c}{\textbf{Inferred}} &
\multicolumn{2}{c}{\textbf{Multiple}} &
\multicolumn{1}{c}{\multirow{2}{*}{\textbf{Overall}}} \\
\cmidrule(lr){2-5}\cmidrule(lr){6-7}\cmidrule(lr){8-9}
 & SN & LN & IN & CN & PS & II & TW & MS & \\
\midrule
GPT-5.6-Luna & \textbf{81.3} & \textbf{5.0} & 16.0 & 3.3 & 26.7 & 10.0 & 1.7 & 1.7 & 21.3 \\
GPT-5.5 & \underline{75.0} & 0.0 & 20.0 & 0.0 & \underline{35.0} & 11.7 & \textbf{3.3} & 0.0 & 20.8 \\
GPT-5.4 & 15.0 & 1.3 & 20.0 & 0.0 & 15.0 & 11.7 & 1.7 & 0.0 & 8.3 \\
GPT-5.2 & 25.0 & 0.0 & 28.0 & 0.0 & 8.3 & \textbf{15.0} & 1.7 & \textbf{3.3} & 10.6 \\
Claude-Opus-5 & 48.8 & 2.5 & \underline{30.0} & 3.3 & 30.0 & 11.7 & \textbf{3.3} & \textbf{3.3} & 17.9 \\
Claude-Opus-4.6 & \underline{75.0} & 1.3 & 28.0 & \textbf{6.7} & \textbf{36.7} & \textbf{15.0} & 1.7 & 1.7 & \textbf{22.9} \\
Gemini-3.8-Flash & 45.0 & 0.0 & 14.0 & 0.0 & 23.3 & 5.0 & 0.0 & 0.0 & 12.5 \\
Gemini-3.6-Flash & 23.8 & 0.0 & 6.0 & 0.0 & 23.3 & 3.3 & 0.0 & 1.7 & 8.1 \\
Gemini-3.1-Pro & 31.3 & 0.0 & 8.0 & 0.0 & 18.3 & 6.7 & 0.0 & 0.0 & 9.2 \\
Doubao-Seed-2.0-Pro & 26.3 & 1.3 & 6.0 & 0.0 & 15.0 & 10.0 & 0.0 & 0.0 & 8.3 \\
GLM-5V-Turbo & 50.0 & 0.0 & 18.0 & 0.0 & 25.0 & \textbf{15.0} & 0.0 & 0.0 & 15.2 \\
Kimi-K3 & 67.5 & \underline{3.8} & \textbf{42.0} & \textbf{6.7} & 31.7 & \textbf{15.0} & 0.0 & 0.0 & \underline{22.5} \\
\bottomrule
\end{tabular*}
\end{table}

When the destination is provided explicitly and remains visible in short-range goal navigation tasks, some models achieve meaningful success on short routes, showing that current agents can combine local spatial perception with action selection when progress can be judged from the immediate scene.


However, navigation breaks down once the route extends across several streets. 
Nearly every completed model fails to reach the endpoint when the same point-to-point task is evaluated at a longer scale.
\textbf{Longer routes require agents to connect successive local decisions to a feasible path through the city.} 
They must also revise their movements when an initially plausible approach becomes blocked. 
Strong local perception does not reliably translate into long-range navigation under our evaluation protocol. 
In the analyzed trajectories, agents repeat blocked actions or lose earlier progress, without consistently advancing toward the goal.

Instruction-dependent navigation exposes another failure that cannot be explained by travel distance alone.
Routes that require the agent to infer the intended destination are comparable in length to short-range navigation, yet performance falls substantially.
The agent must convert the user's request into a stable goal and continue following that goal as the observation changes.
The decline shows that instruction understanding and adherence remain unreliable during interactive exploration.

\begin{figure}[t!]
\centering
\begin{minipage}[t]{0.498\textwidth}
\vspace{0pt}
\centering
\includegraphics[width=\linewidth]{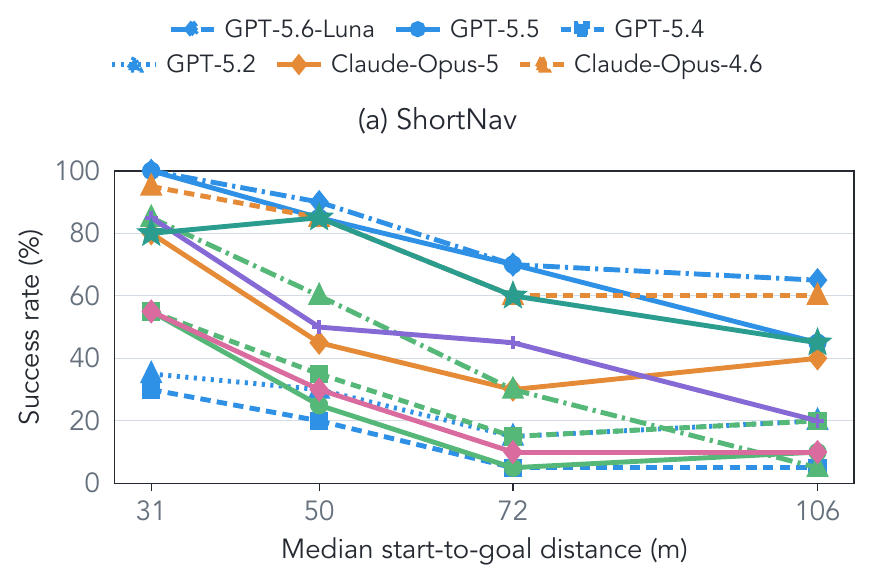}
\end{minipage}\hfill
\begin{minipage}[t]{0.498\textwidth}
\vspace{0pt}
\centering
\includegraphics[width=\linewidth]{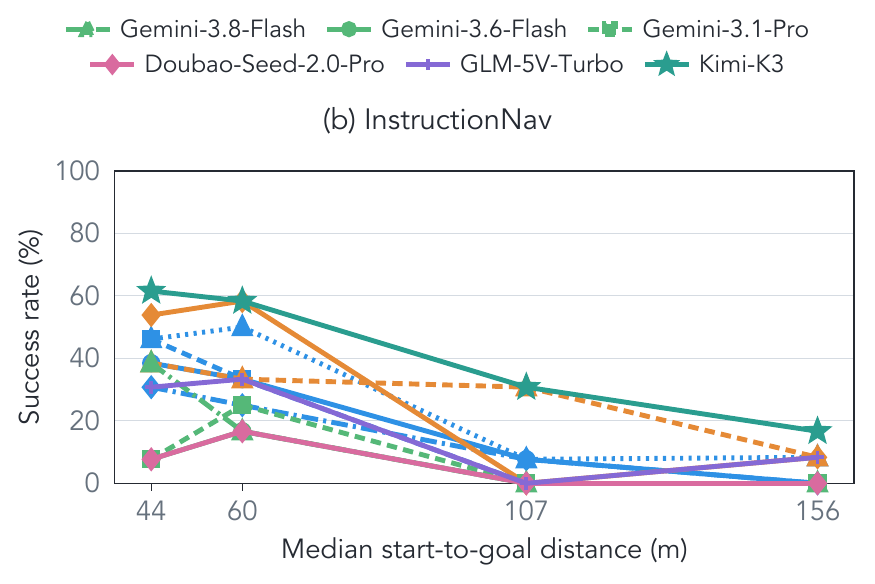}
\end{minipage}
\caption{Navigation success as a function of interaction horizon. The panels report success across four equal-count bins for (a) ShortNav and (b) InstructionNav.}
\label{fig:rq2-navigation-horizon}
\end{figure}

We further conduct a distance-stratified navigation-horizon analysis by partitioning ShortNav and InstructionNav episodes into four equal-count bins within each task family according to the initial straight-line start-to-goal distance.
Figure~\ref{fig:rq2-navigation-horizon} reports the corresponding bin-wise success rates against the median start-to-goal distance of each bin.
Across both task families, success is concentrated in the shorter-distance bins and generally declines as the navigation horizon increases.
This shows that the failure is not confined to a discrete ShortNav--LongNav split. 
Even within nominally local tasks, increasing the interaction horizon systematically weakens closed-loop execution.

\textbf{Interestingly, interactive navigation magnifies model differences that appear modest in question answering}.
The current models remain relatively close on the local QA tasks, while navigation produces a much wider separation.
Even within the same GPT family, GPT-5.5 and GPT-5.4 diverge sharply on short routes.
Every action changes the next observation and decision context, so small differences in spatial recovery can compound over an episode.
Navigation in the real-scale embodied environment exposes the stability of closed-loop behavior that isolated question answering does not reveal.
\label{app:rq2-partial-progress}

\begin{figure}[!htbp]
\centering
\begin{minipage}[t]{0.495\textwidth}
\vspace{0pt}
\centering
\includegraphics[width=\linewidth]{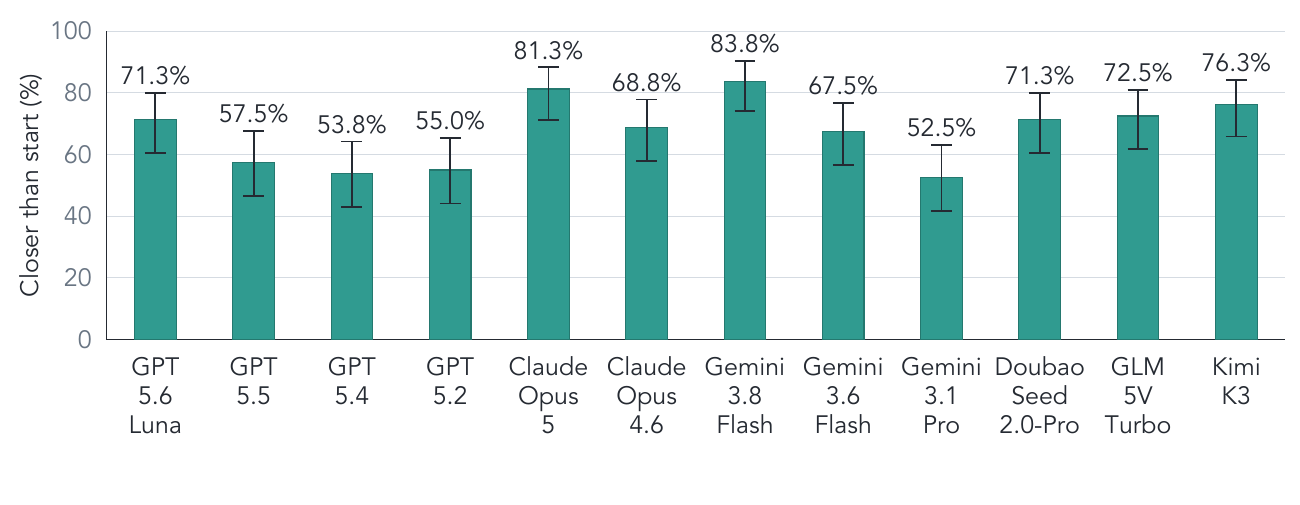}
\end{minipage}\hfill
\begin{minipage}[t]{0.495\textwidth}
\vspace{0pt}
\centering
\includegraphics[width=\linewidth]{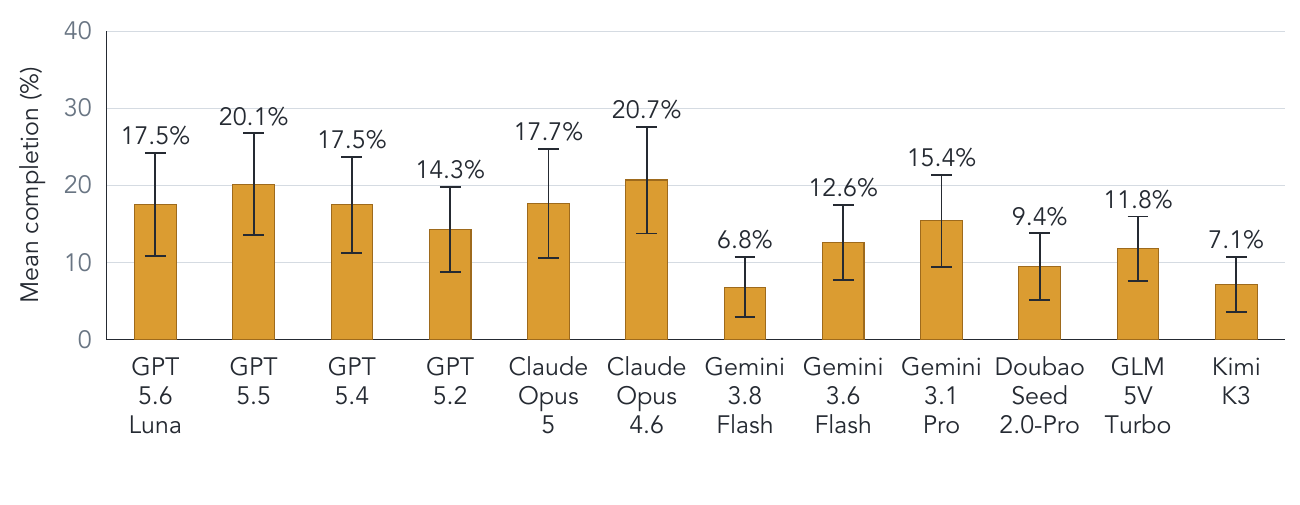}
\end{minipage}
\caption{Long-range navigation progress and multi-stop completion. The left panel reports the proportion of LongNav episodes that end closer to the goal than at initialization. The right panel reports the mean proportion of required destinations reached in multi-stop planning. Error bars show 95\% confidence intervals.}
\label{fig:rq2-partial-progress}
\end{figure}

We further add two trajectory-level analyses.
For long-range navigation, we measure the fraction of episodes that end closer to the goal than they begin. 
For multi-stop planning, we measure the mean fraction of required destinations reached.
Figure~\ref{fig:rq2-partial-progress} shows that more than 50\% of long-range episodes still reduce the goal distance, despite the near-zero full-task success in Table~\ref{tab:rq2-success}.
Multi-stop agents likewise reach 6.8--20.7\% of their required destinations on average while rarely completing the entire plan.
The agents often establish a useful local direction, but fail to preserve the global route state long enough to finish.
Further analysis shows that LongNav trajectories often lose earlier progress, while failures span premature stopping, absent progress, lost progress, and retained but incomplete progress (Appendix~\ref{app:longnav-checkpoint}~and~\ref{app:longnav-failure-types}).

\label{app:longnav-case-studies}

\begin{figure}[t!]
\centering
\casepanel{RQTwo}{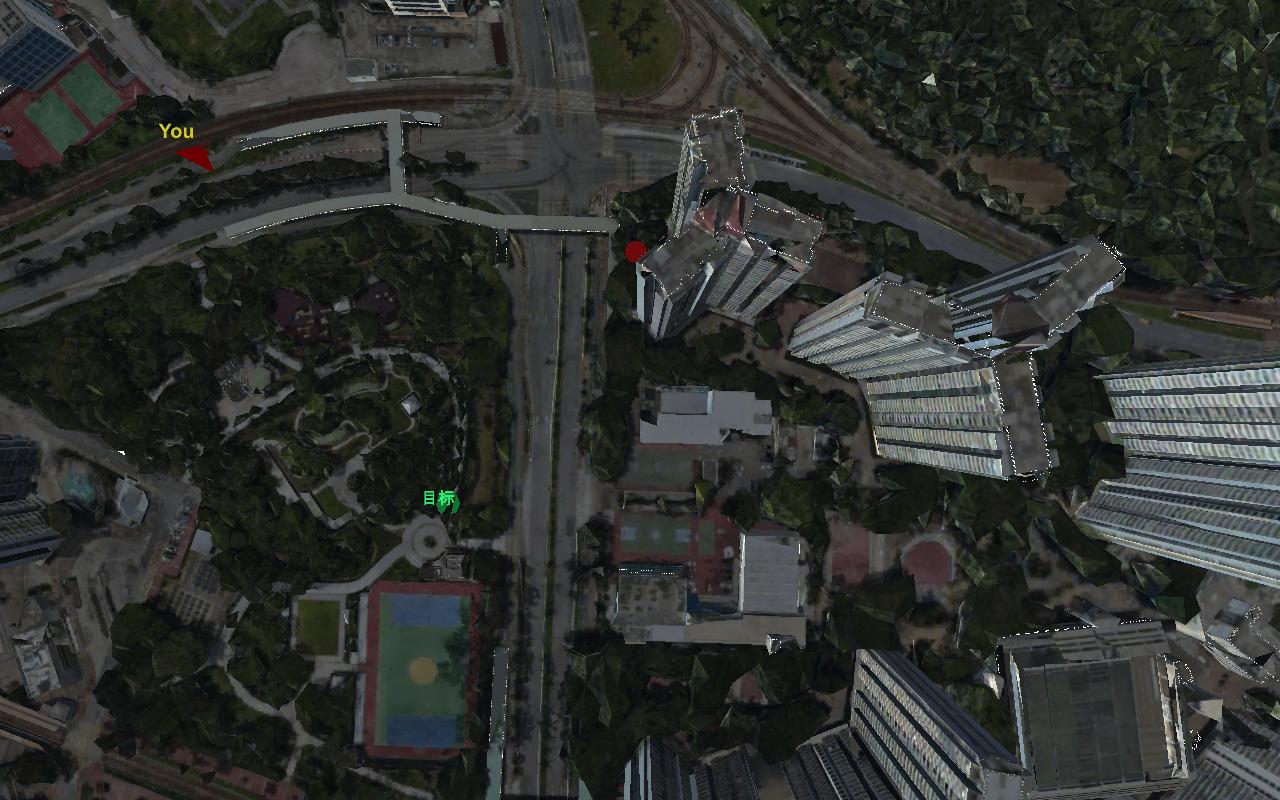}{a}{29}{close map}\hfill
\casepanel{RQTwo}{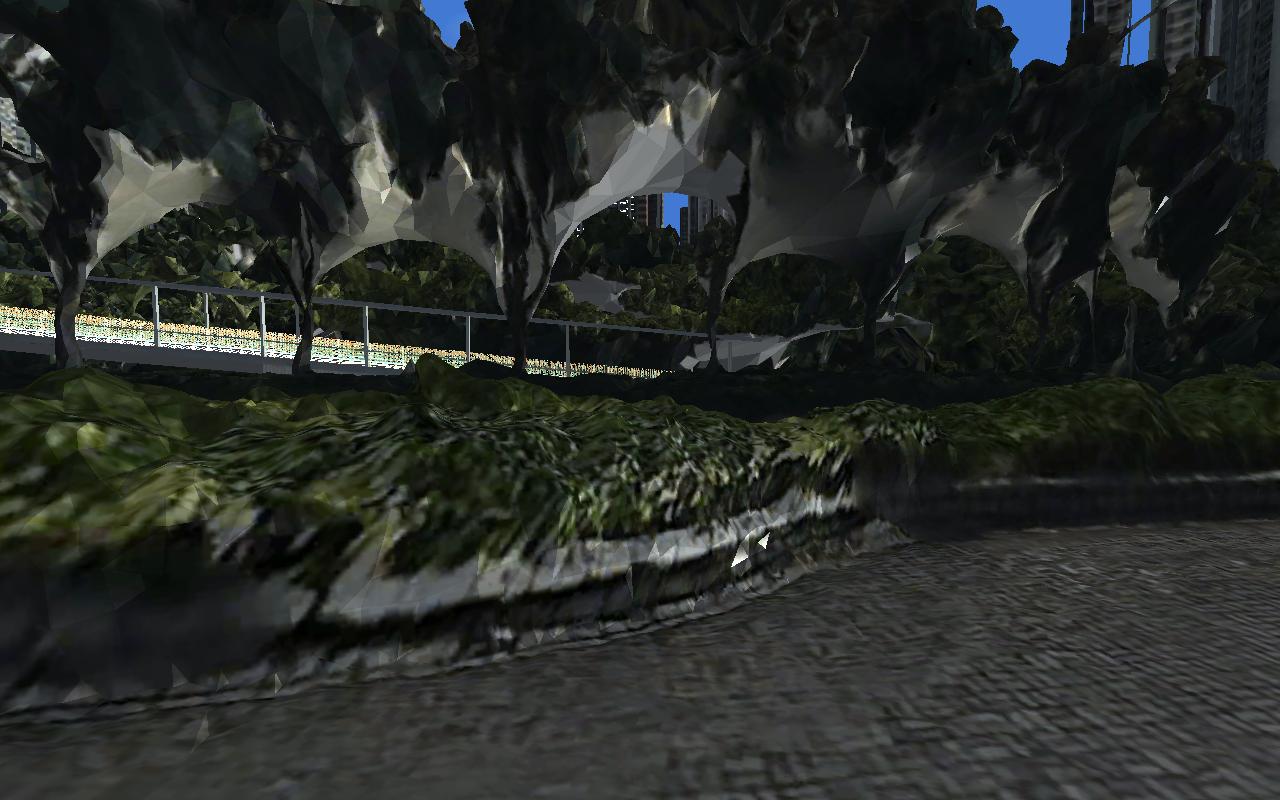}{b}{34}{look left $90^\circ$, pitch 0}\hfill
\casepanel{RQTwo}{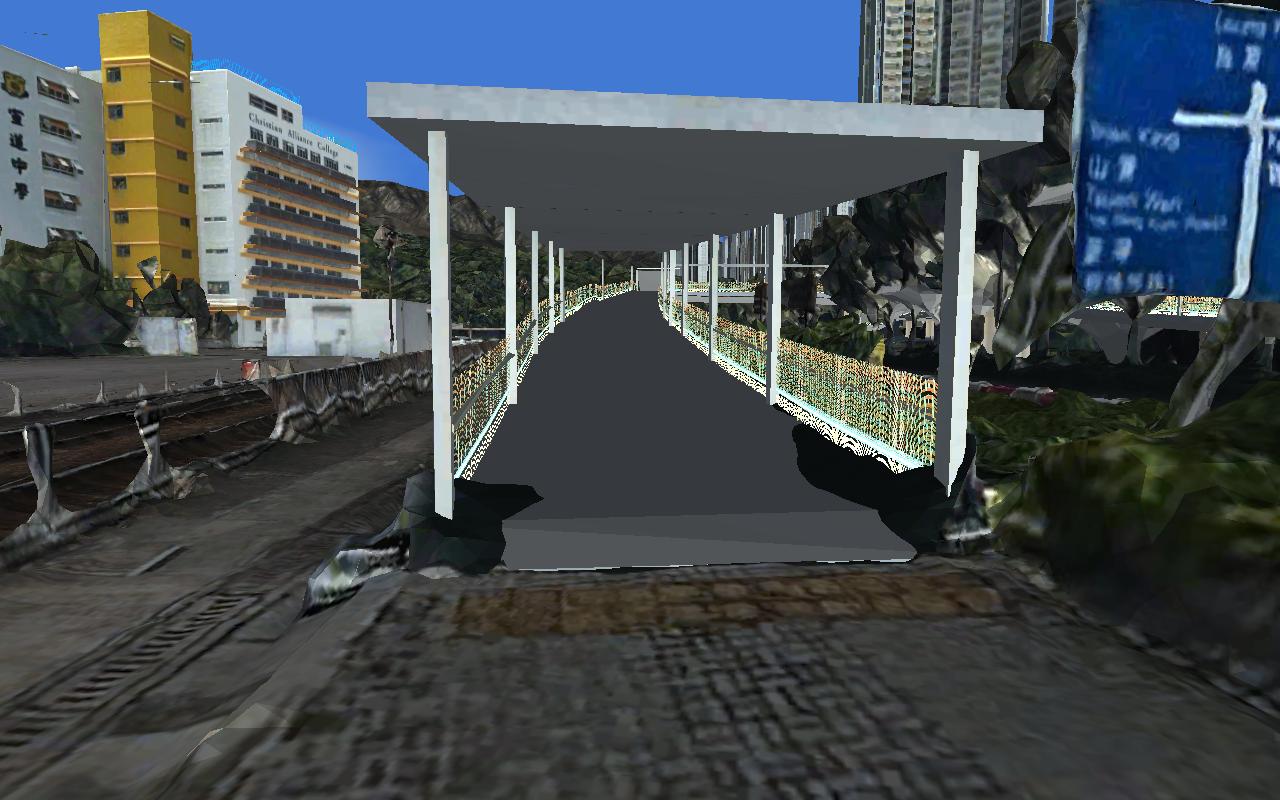}{c}{35}{sprint forward 2.0 s}\hfill
\casepanel{RQTwo}{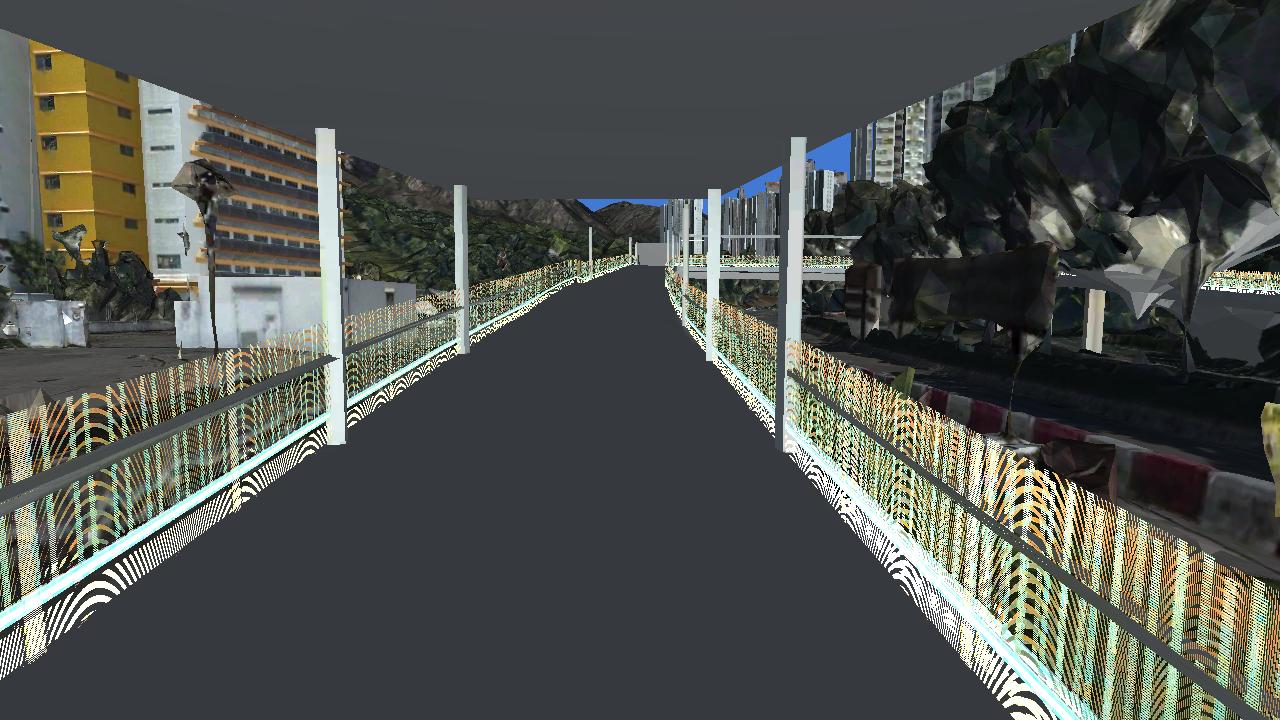}{d}{36}{sprint forward 2.0 s}
\par\smallskip
\casepanel{RQTwo}{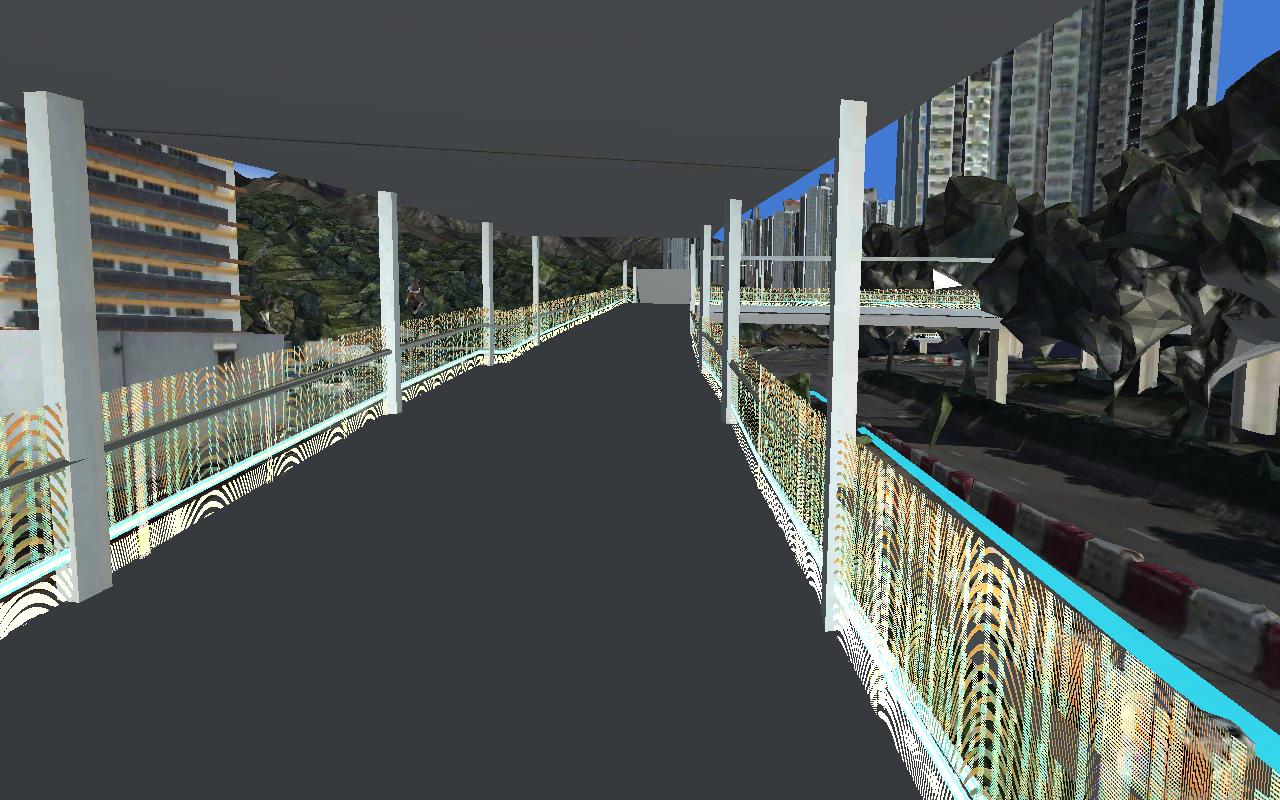}{e}{37}{sprint forward 2.0 s}\hfill
\casepanel{RQTwo}{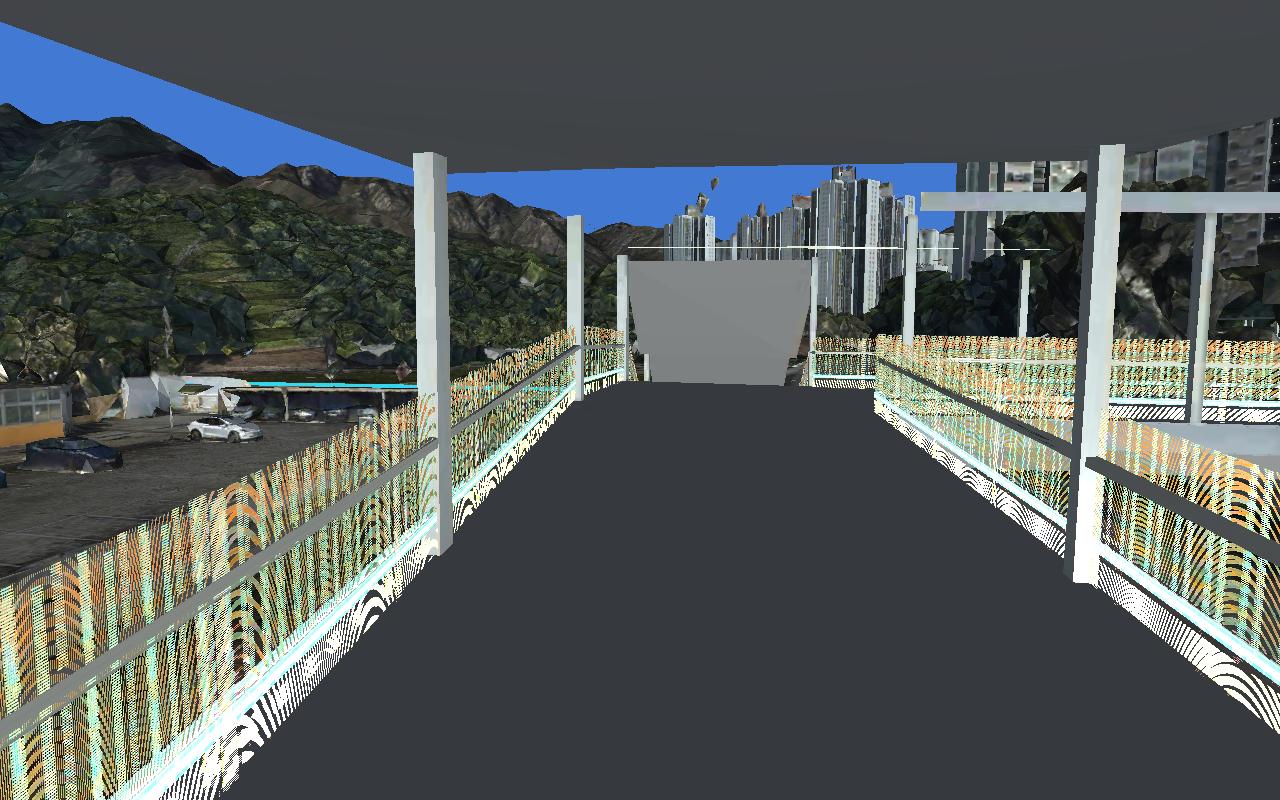}{f}{39}{sprint forward 2.0 s}\hfill
\casepanel{RQTwo}{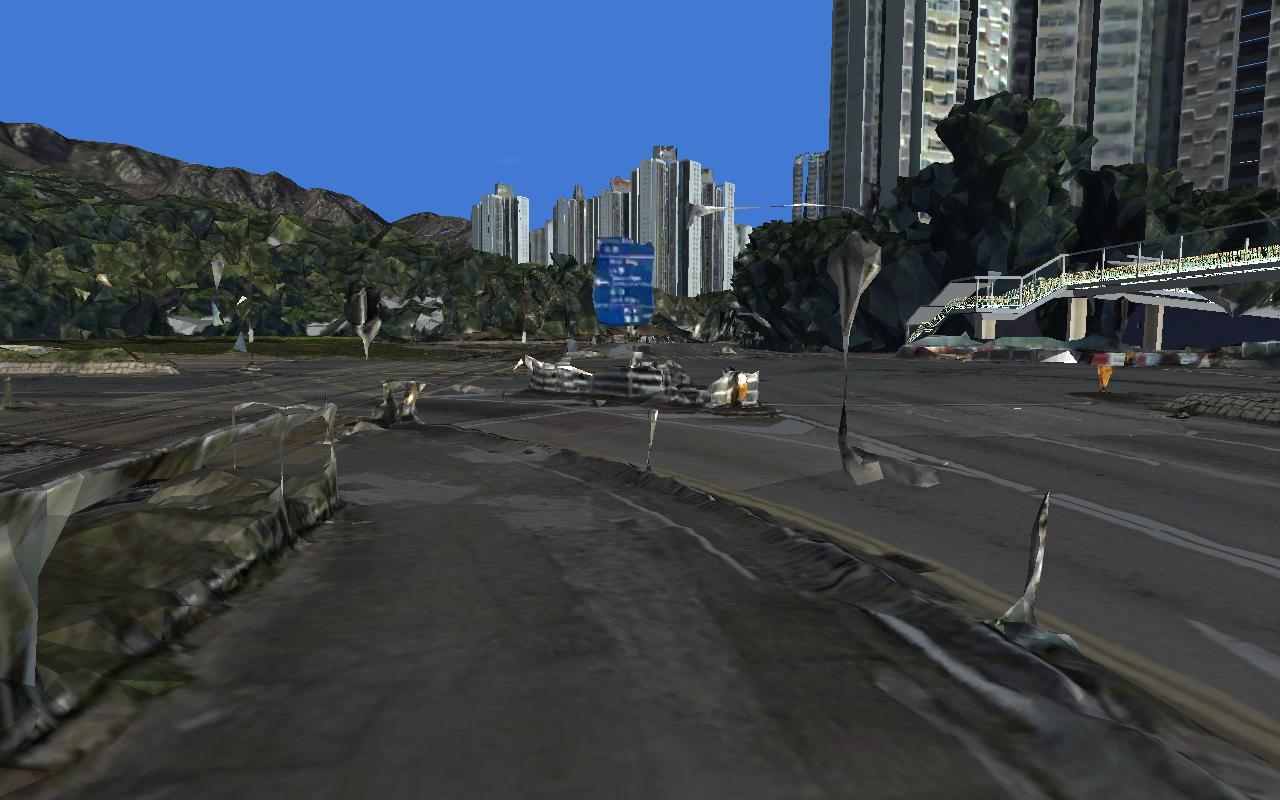}{g}{43}{open map}\hfill
\casepanel{RQTwo}{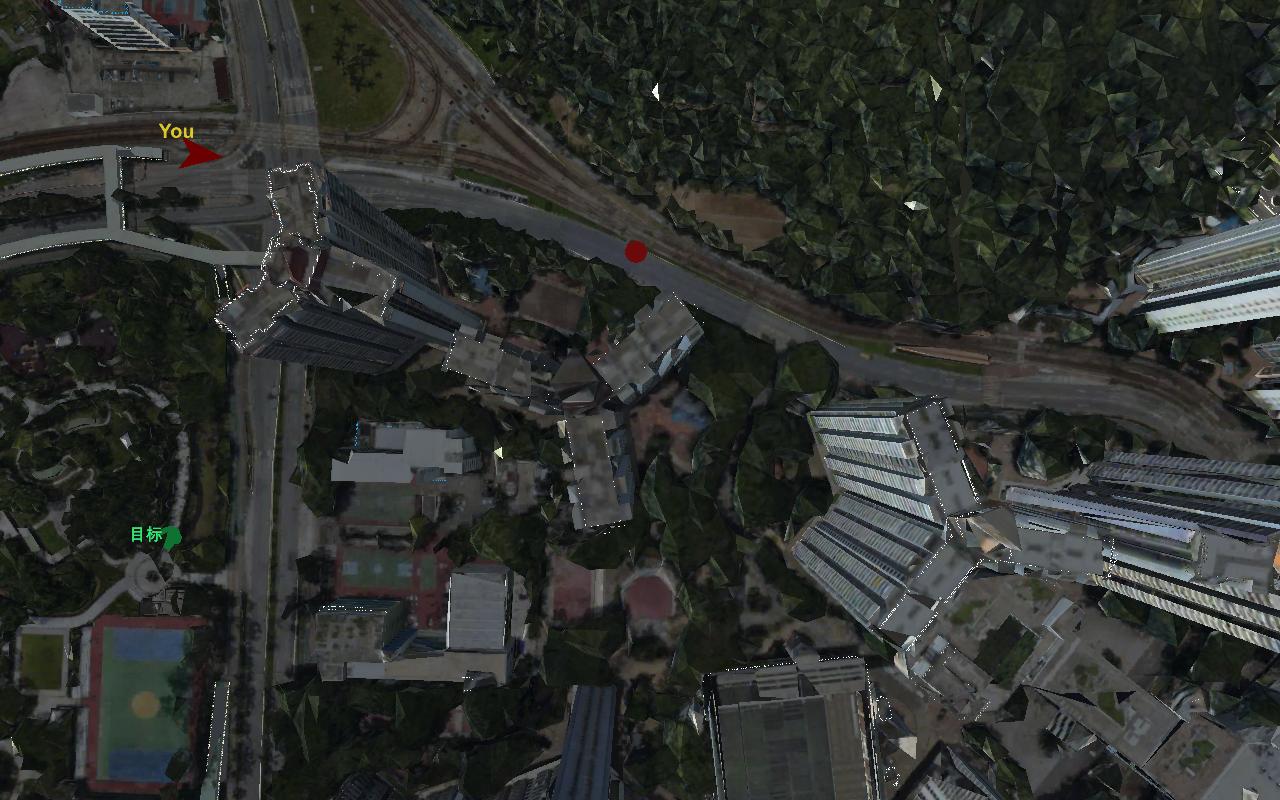}{h}{44}{close map}
\caption{Example of GPT-5.5 traversing a complex interchange using the official footbridge and continuing steadily toward the goal.}
\label{fig:rq2-bridge-case}
\end{figure}

\begin{figure}[t!]
\centering
\casepanel{RQTwo}{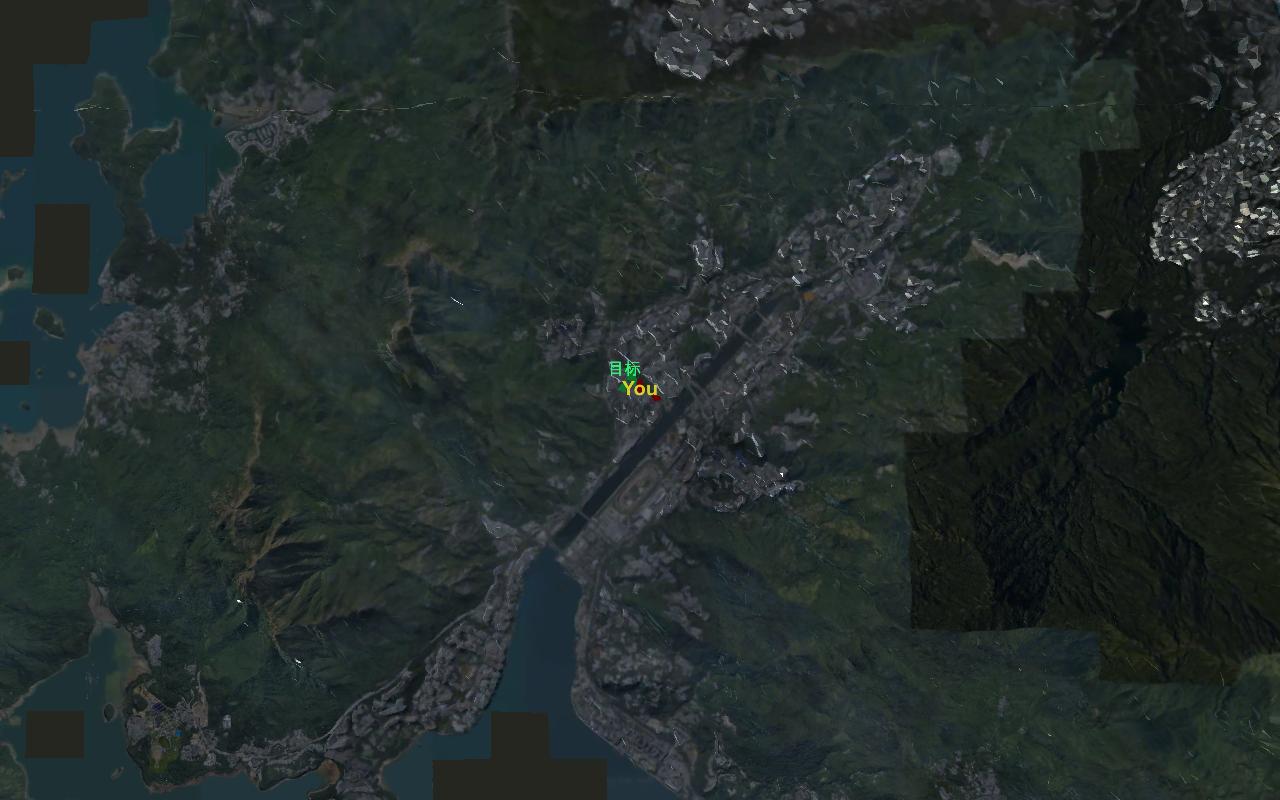}{a}{2}{zoom map by $0.25\times$}\hfill
\casepanel{RQTwo}{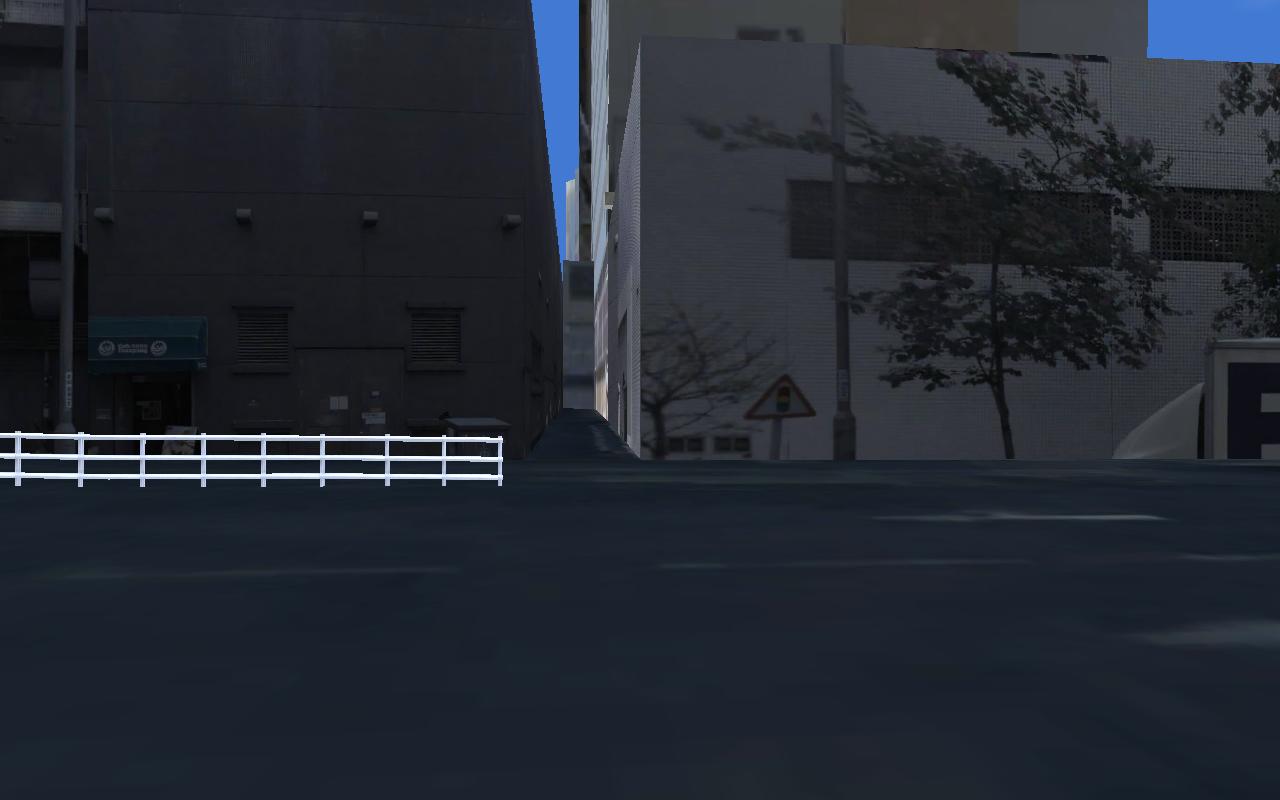}{b}{6}{sprint forward 2.0 s}\hfill
\casepanel{RQTwo}{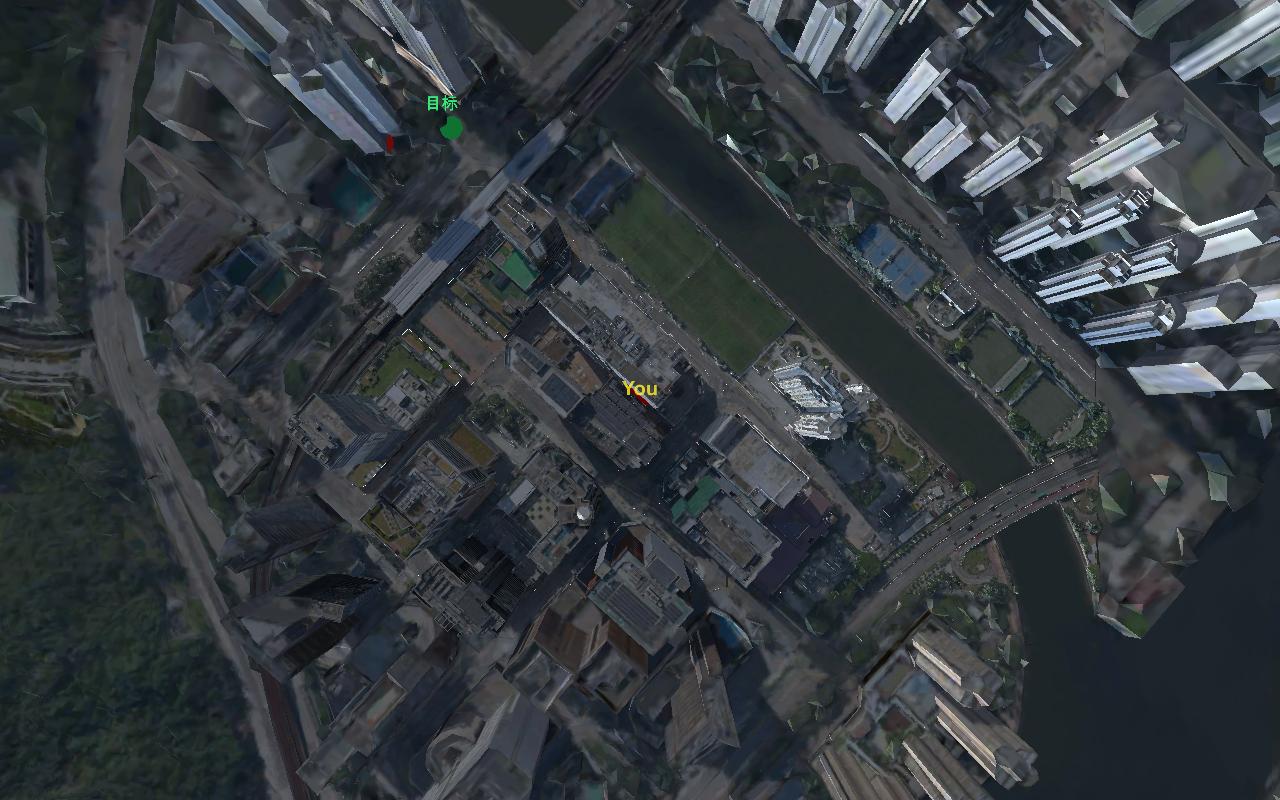}{c}{12}{close map}\hfill
\casepanel{RQTwo}{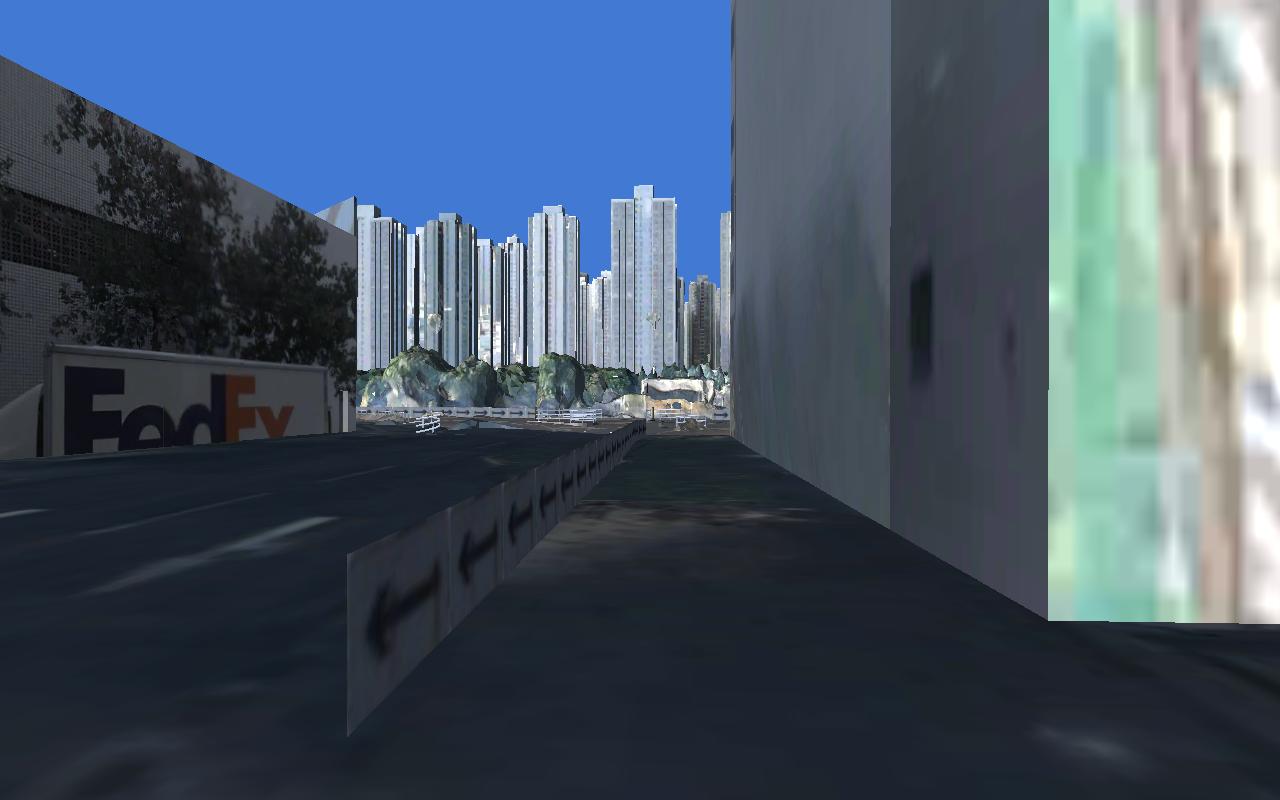}{d}{44}{sprint forward 2.0 s}
\par\smallskip
\casepanel{RQTwo}{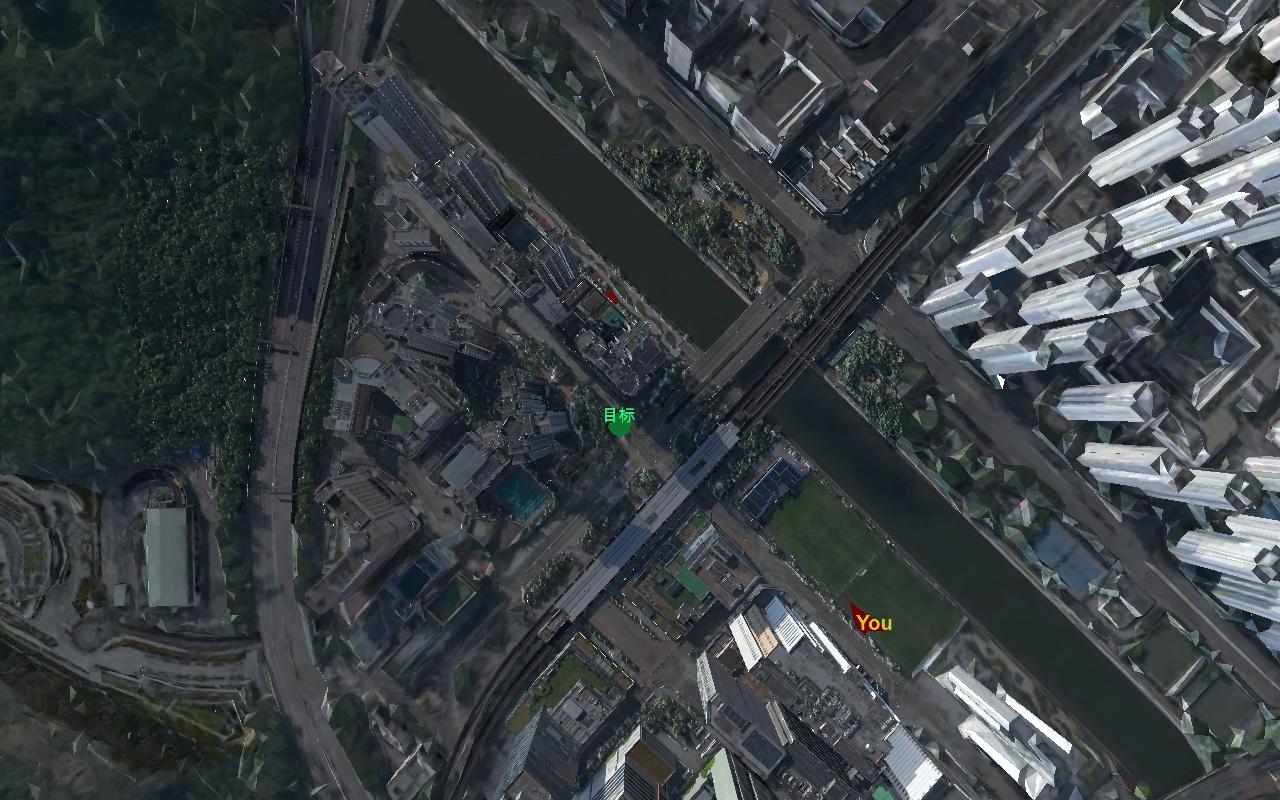}{e}{59}{close map}\hfill
\casepanel{RQTwo}{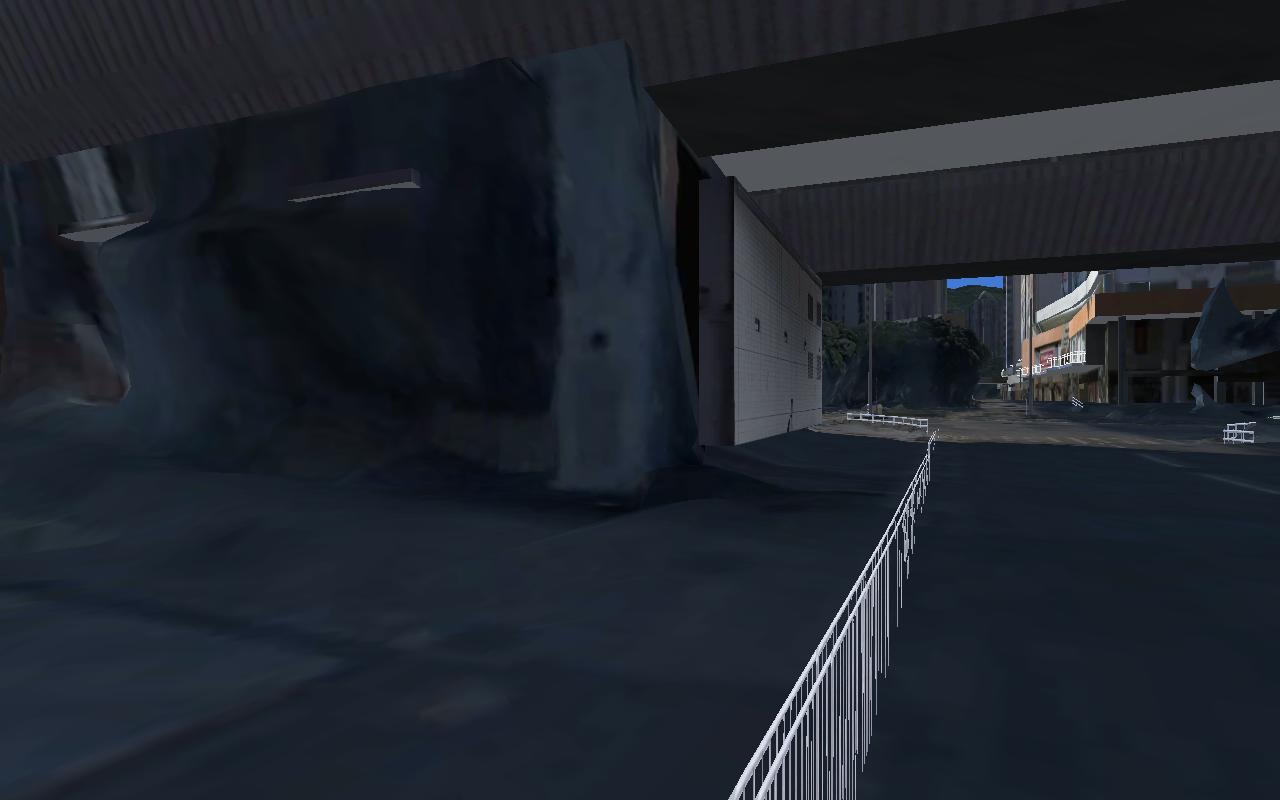}{f}{73}{sprint forward 2.0 s, yaw $-1$}\hfill
\casepanel{RQTwo}{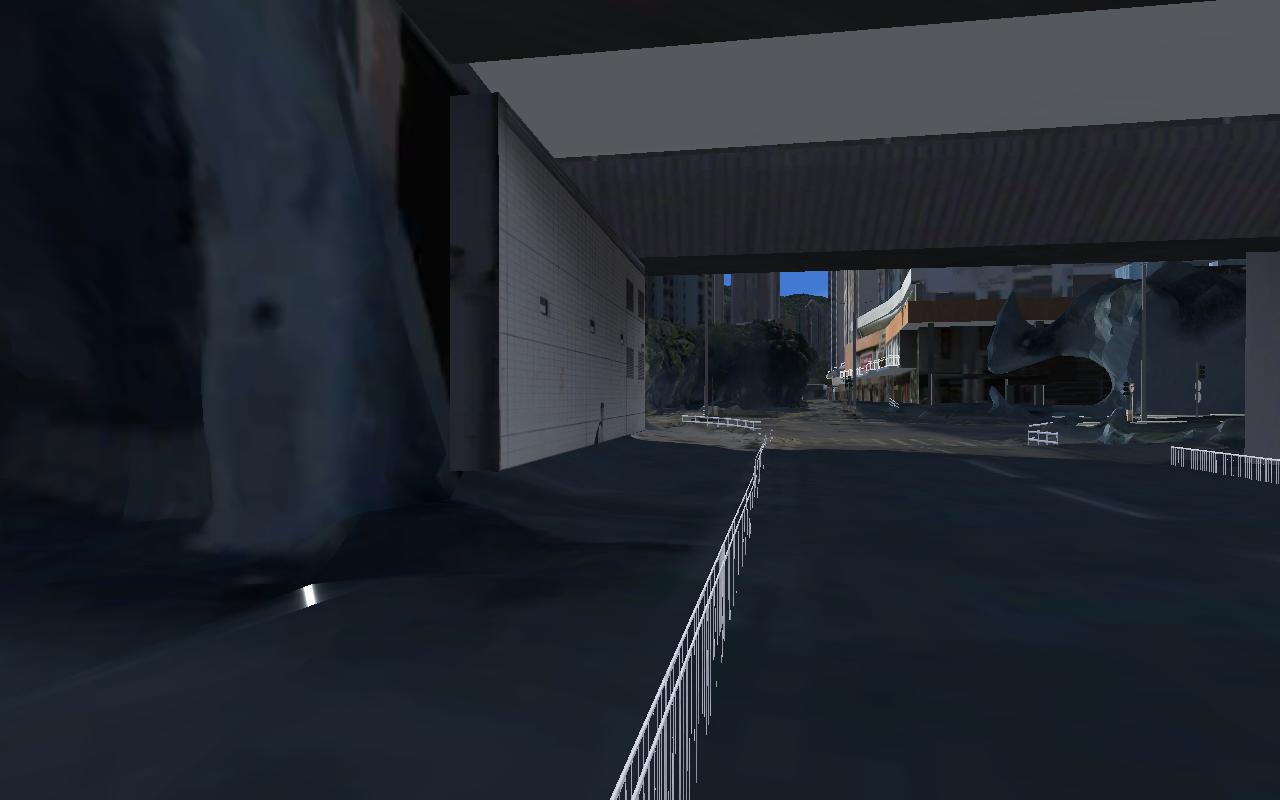}{g}{81}{sprint forward 1.5 s, yaw $-1$}\hfill
\casepanel{RQTwo}{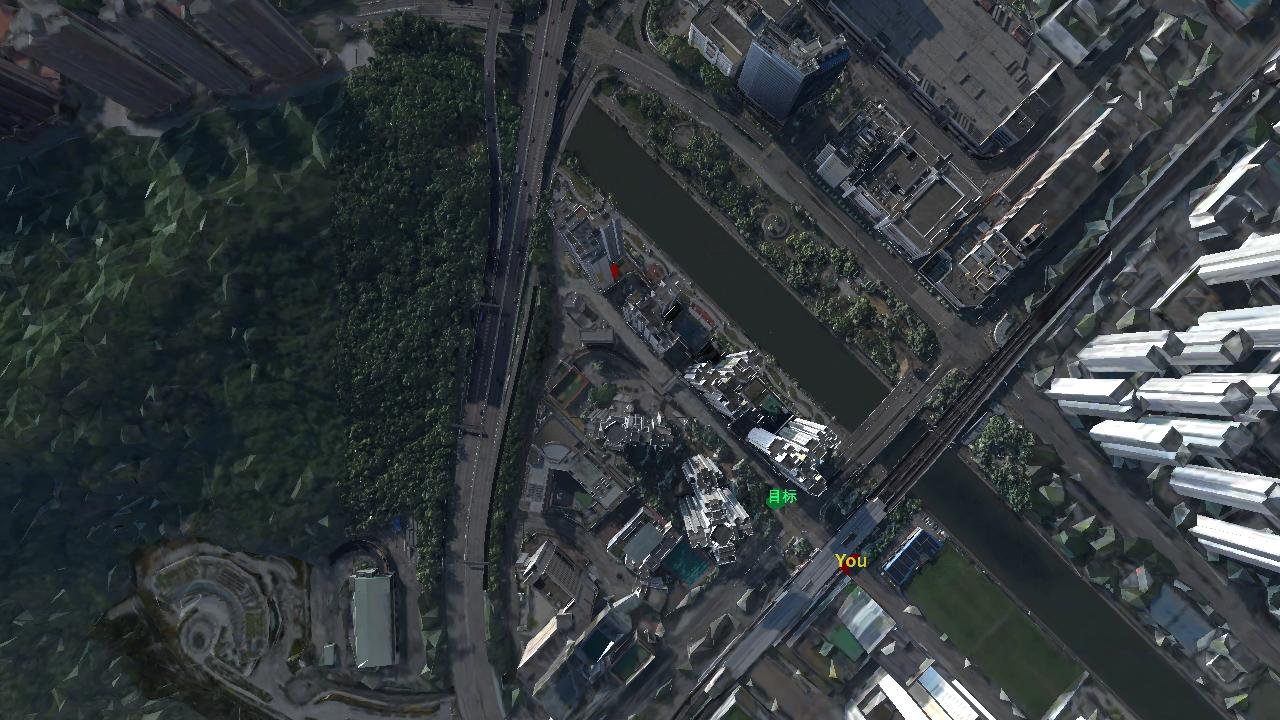}{h}{94}{close map}
\caption{Example of GPT-5.5 crossing the road toward the goal direction before becoming blocked by a central obstacle and being unable to continue.}
\label{fig:rq2-failure-case}
\end{figure}

Figures~\ref{fig:rq2-bridge-case} and~\ref{fig:rq2-failure-case} present two attempts by GPT-5.5 to move toward the goal during long-range exploration.
In Figure~\ref{fig:rq2-bridge-case}, a hedge blocks the direct continuation, and the agent identifies the official ramp, traverses the complex footbridge, and continues steadily toward the goal.
This attempt uses the local pedestrian structure correctly even though the full long-range episode remains unfinished.
In Figure~\ref{fig:rq2-failure-case}, the agent tries to reduce the apparent distance to the goal by crossing the road in its estimated goal direction.
It becomes blocked by an obstacle in the middle of the road and repeats forward actions without revising the route.
The contrast shows that long-range exploration depends on distinguishing geometric proximity from a traversable path and correcting an attempt when the local scene invalidates it.

\rqsection{RQThree}{RQ3: How do changes in the city affect grounded perception and action?}
\label{sec: RQThree}

We finally study whether the capabilities observed in the default city persist across dynamic changes.
We evaluate every question-answering task from RQ1 and the short-range navigation task under matched conditions with various weather and time.
Table~\ref{tab:rq3-weather} keeps the location and task instance fixed while changing clear daytime and weather conditions.
\begin{table}[t!]
\centering
\caption{Local question-answering accuracy and short-navigation success compare robustness across different daytime and weather conditions.}
\label{tab:rq3-weather}
\setlength{\tabcolsep}{4.1pt}
\renewcommand{\arraystretch}{1.12}
\resizebox{\textwidth}{!}{%
\begin{tabular}{l*{5}{c}@{\hspace{11pt}}*{5}{c}}
\toprule
\multirow{2}{*}{\textbf{Model}} &
\multicolumn{5}{c}{\textbf{Local QA Accuracy}} &
\multicolumn{5}{c}{\textbf{Short Navigation Success}} \\
\cmidrule(lr){2-6}\cmidrule(lr){7-11}
 & Clear & Dusk & Night & Cloudy & Rain & Clear & Dusk & Night & Cloudy & Rain \\
\midrule
GPT-5.6-Luna & 63.2 & 56.4 & 60.9 & 62.3 & 57.7 & \textbf{81.3} & \textbf{73.8} & \underline{71.3} & \textbf{77.5} & \textbf{76.3} \\
GPT-5.5 & 63.6 & 59.1 & 61.4 & 62.7 & 61.4 & \underline{75.0} & \underline{72.5} & \textbf{75.0} & \underline{76.3} & \underline{72.5} \\
GPT-5.4 & 56.8 & 44.5 & 48.6 & 58.2 & 53.6 & 15.0 & 13.8 & 18.8 & 11.3 & 16.3 \\
GPT-5.2 & 55.0 & 44.1 & 48.6 & 55.9 & 53.6 & 25.0 & 26.3 & 17.5 & 27.5 & 21.3 \\
Claude-Opus-5 & \textbf{79.1} & 66.4 & 69.5 & 75.5 & 70.5 & 48.8 & 46.3 & 41.3 & 51.3 & 46.3 \\
Claude-Opus-4.6 & 70.5 & 55.5 & 63.2 & 69.1 & 67.7 & \underline{75.0} & 70.0 & 65.0 & 73.8 & 65.0 \\
Gemini-3.8-Flash & \underline{78.6} & \textbf{74.1} & \textbf{76.4} & \textbf{75.9} & \underline{72.3} & 45.0 & 46.3 & 48.8 & 46.3 & 41.3 \\
Gemini-3.6-Flash & 77.7 & \underline{68.6} & \underline{72.7} & 71.4 & \textbf{73.2} & 23.8 & 22.5 & 26.3 & 18.8 & 12.5 \\
Gemini-3.1-Pro & 53.2 & 42.3 & 37.3 & 46.8 & 45.5 & 31.3 & 28.8 & 31.3 & 30.0 & 32.5 \\
Doubao-Seed-2.0-Pro & 64.1 & 57.7 & 60.9 & 66.8 & 64.1 & 26.3 & 16.3 & 20.0 & 28.8 & 25.0 \\
GLM-5V-Turbo & 58.2 & 46.4 & 47.7 & 56.8 & 55.0 & 50.0 & 47.5 & 40.0 & 47.5 & 45.0 \\
Kimi-K3 & 76.4 & 65.0 & 62.7 & \textbf{75.9} & \underline{72.3} & 67.5 & 63.8 & 68.8 & 70.0 & 66.3 \\
\bottomrule
\end{tabular}}
\end{table}

\textbf{Dynamic changes in weather and time can weaken the ability of MLLM agents to answer urban questions reliably.}
Dusk and night reduce the visibility of signs, storefronts, and other local evidence, which increases the difficulty of grounding an answer in the scene.
Robustness is not uniform even within a model family.
GPT-5.6-Luna and Gemini-3.8-Flash remain comparatively stable across the matched conditions, while GPT-5.2 and Gemini-3.1-Pro change much more as the scene moves from daylight into lower-visibility periods.
This contrast shows that strong performance in the default condition does not ensure stable spatial reasoning when illumination changes.

The effect of weather and time is less systematic on short-range navigation than on question answering.
Conditions that reduce local QA accuracy do not consistently reduce navigation success, and the ordering of conditions changes across models.
They show that appearance changes alone cannot explain the navigation failures, which also depend on route-state maintenance and action execution. 
Interestingly, although visibility is lower at night, agents generally perform worse under dusk conditions than at night, possibly due to interference from the lighting conditions at dusk.

Beyond appearance changes, we next alter the navigable world to test whether agents can adapt their actions and routes to changing environments.
The road-closure task invalidates a pedestrian segment after the route has begun, while navigation among pedestrians introduces moving obstacles around a fixed destination.
Table~\ref{tab:rq3-dynamics} reports navigation success and pedestrian-network adherence alongside closure compliance rate under road closure and pedestrian collisions.
\begin{table}[t!]
\centering
\caption{Dynamic-environment results, where SR is the goal-reaching rate, PNA is the fraction of action time spent on the pedestrian network, CCR is the closure compliance rate that represents the fraction of road-closure episodes that respect the closure, and PCR is the pedestrian-collision rate.}
\label{tab:rq3-dynamics}
\setlength{\tabcolsep}{3.0pt}
\renewcommand{\arraystretch}{1.12}
\begin{tabular*}{0.8\textwidth}{@{\extracolsep{\fill}}l*{6}{>{\centering\arraybackslash}p{3.2em}}@{}}
\toprule
\multirow{2}{*}{\textbf{Model}} &
\multicolumn{3}{c}{\textbf{Road Closure}} &
\multicolumn{3}{c}{\textbf{Pedestrians}} \\
\cmidrule(lr){2-4}\cmidrule(lr){5-7}
 & SR$\uparrow$ & PNA$\uparrow$ & CCR$\uparrow$ & SR$\uparrow$ & PNA$\uparrow$ & PCR$\downarrow$ \\
\midrule
GPT-5.6-Luna & 0.0 & \textbf{98.4} & 20.0 & 0.0 & \textbf{98.6} & 82.5 \\
GPT-5.5 & 0.0 & 94.7 & 13.3 & 0.0 & 93.8 & 83.8 \\
GPT-5.4 & 0.0 & \underline{96.4} & 16.7 & 0.0 & \underline{98.5} & 80.0 \\
GPT-5.2 & 0.0 & 94.2 & 16.7 & 1.3 & 95.4 & \underline{78.8} \\
Claude-Opus-5 & 0.0 & 93.6 & \textbf{46.7} & \underline{2.5} & 94.4 & 87.5 \\
Claude-Opus-4.6 & \textbf{3.3} & 93.0 & 26.7 & \textbf{3.8} & 92.9 & 80.0 \\
Gemini-3.8-Flash & 0.0 & 95.8 & 20.0 & 0.0 & 97.5 & 85.0 \\
Gemini-3.6-Flash & 0.0 & 95.4 & 16.7 & 0.0 & 96.0 & 90.0 \\
Gemini-3.1-Pro & 0.0 & 93.4 & 16.7 & 0.0 & 94.0 & 88.8 \\
Doubao-Seed-2.0-Pro & 0.0 & 93.5 & 10.0 & 0.0 & 95.0 & \textbf{76.3} \\
GLM-5V-Turbo & 0.0 & 93.3 & 20.0 & 0.0 & 95.9 & 82.5 \\
Kimi-K3 & \textbf{3.3} & 95.1 & \underline{40.0} & \underline{2.5} & 90.6 & \underline{78.8} \\
\bottomrule
\end{tabular*}
\end{table}

Closure Compliance Rate remains low across models even as PNA stays high, showing that agents often continue to produce locally compliant movement without recovering safe goal-directed progress when the available route is altered.
PCR is high for every model under moving pedestrians, so remaining on the pedestrian network does not imply collision-aware control.
These measures expose failures in route revision and local motion adaptation within the changed scenes.

The same separation recurs at each spatial scale.
Models recognize local evidence from the current viewpoint more reliably than they orient it, execute useful local motion more reliably than they complete a route, and preserve compliant movement more reliably than they update a plan after the city changes.
Across the experiments, the grounded spatial state ceases to guide action toward the objective long before the agent stops producing plausible local movements.
\section{Conclusion}

In this paper, we investigate whether current MLLM agents can carry local spatial understanding into sustained action in a complicated real-scale city. 
We introduced \UrbanGroundName{}, a georegistered replica of Hong Kong that supports closed-loop interaction under physical change. 
Our study follows the growth of the spatial problem from active question answering to navigation across longer distances and less explicit instructions, then to adaptation when the city changes. 
The evaluation shows that current agents already possess useful atomic abilities in visual recognition and short-range spatial reasoning. 
However, failure usually emerges over longer routes. 
Local abilities do not compose into sustained exploration, so small errors accumulate without effective correction. 
Changes in visibility alter question answering, while road closures and moving pedestrians expose weak online updating and constraint adherence. 
These findings show that city-scale agency depends on maintaining a spatial estimate after the current view disappears, then revising it in response to new evidence whenever earlier plans cease to be feasible. 
We hope \UrbanGroundName{} supports future research on agents capable of moving through complex cities with greater adaptability.

\section*{AI Use Statement}

We used AI assistants to help polish the manuscript. We also used MLLM APIs in our experiments to study the capabilities of agents in an open urban environment. We did not use AI for any other purpose. We take responsibility for the final content of this work.

\section*{Ethics Statement}

We study MLLM agents in an urban simulator built from publicly available geospatial data from Hong Kong. Human annotation is limited to constructing and verifying the benchmark tasks. Since all experiments take place within the simulator and do not involve real-world deployment, we do not foresee ethical concerns in this paper.

\section*{Reproducibility Statement}

We describe our experimental setup and evaluation metrics in Section~\ref{sec: Experiments}. The framework implementation is detailed in Section~\ref{app:framework-implementation}. We explain task construction and human verification in Appendix~\ref{sec: Experimental Task Design} and Section~\ref{app:human-annotation}. We also provide the prompts used in our experiments in Appendix~\ref{app:prompts} and the complete results in Appendix~\ref{app:full-results}.

%% file: sections/full_results_appendix.tex
\section{Complete Experimental Results}
\label{app:full-results}

The main text reports compact tables organized by the three research questions.
The tables below retain every evaluated model and report all task-level results as percentages.

\newcommand{\fullmodelrows}{%
GPT-5.6-Luna & 78.8 & 48.3 & 58.8 & \textbf{81.3} & \textbf{5.0} & 16.0 & 3.3 & 26.7 & 10.0 & 1.7 & 1.7 & 0.0 & 0.0 \\
GPT-5.5 & 82.5 & 40.0 & 62.5 & \underline{75.0} & 0.0 & 20.0 & 0.0 & \underline{35.0} & 11.7 & \textbf{3.3} & 0.0 & 0.0 & 0.0 \\
GPT-5.4 & 75.0 & 31.7 & 57.5 & 15.0 & 1.3 & 20.0 & 0.0 & 15.0 & 11.7 & 1.7 & 0.0 & 0.0 & 0.0 \\
GPT-5.2 & 77.5 & 33.3 & 48.8 & 25.0 & 0.0 & 28.0 & 0.0 & 8.3 & \textbf{15.0} & 1.7 & \textbf{3.3} &0.0 & 1.3 \\
Claude-Opus-5 & 91.3 & \textbf{58.3} & \textbf{82.5} & 48.8 & 2.5 & \underline{30.0} & 3.3 & 30.0 & 11.7 & \textbf{3.3} & \textbf{3.3} & 0.0 & \underline{2.5} \\
Claude-Opus-4.6 & 85.0 & 46.7 & 73.8 & \underline{75.0} & 1.3 & 28.0 & \textbf{6.7} & \textbf{36.7} & \textbf{15.0} & 1.7 & 1.7 & \textbf{3.3} & \textbf{3.8} \\
Gemini-3.8-Flash & \underline{92.5} & \textbf{58.3} & \underline{80.0} & 45.0 & 0.0 & 14.0 & 0.0 & 23.3 & 5.0 & 0.0 & 0.0 & 0.0 & 0.0 \\
Gemini-3.6-Flash & \textbf{93.8} & 56.7 & 77.5 & 23.8 & 0.0 & 6.0 & 0.0 & 23.3 & 3.3 & 0.0 & 1.7 & 0.0 & 0.0 \\
Gemini-3.1-Pro & 82.5 & 23.3 & 46.3 & 31.3 & 0.0 & 8.0 & 0.0 & 18.3 & 6.7 & 0.0 & 0.0 & 0.0 & 0.0 \\
Doubao-Seed-2.0-Pro & 85.0 & 36.7 & 63.8 & 26.3 & 1.3 & 6.0 & 0.0 & 15.0 & 10.0 & 0.0 & 0.0 & 0.0 & 0.0 \\
GLM-5V-Turbo & 80.0 & 36.7 & 52.5 & 50.0 & 0.0 & 18.0 & 0.0 & 25.0 & \textbf{15.0} & 0.0 & 0.0 & 0.0 & 0.0 \\
Kimi-K3 & \underline{92.5} & 55.0 & 76.3 & 67.5 & \underline{3.8} & \textbf{42.0} & \textbf{6.7} & 31.7 & \textbf{15.0} & 0.0 & 0.0 & \textbf{3.3} & \underline{2.5} \\
}

\begin{minipage}{\textwidth}
\centering
\captionsetup{type=table}
\captionof{table}{Complete answer-accuracy and task-success results across the five-level evaluation ladder. Figure~\ref{fig:rq-level-map} shows how the task levels relate to RQ1--RQ3.}
\label{tab:full-success}
\scriptsize
\setlength{\tabcolsep}{2.7pt}
\renewcommand{\arraystretch}{1.07}
\resizebox{\textwidth}{!}{%
\begin{tabular}{l*{13}{c}}
\toprule
\rowcolor{ModelHeadLight}
Model & VR & OU & AEQ & Short & Long & Instr. & Constr. & Search & Intent & Time & Multi & Closure & Ped. \\
\midrule
\fullmodelrows
\bottomrule
\end{tabular}}
\end{minipage}

\vspace{1.4em}

\begin{minipage}{\textwidth}
\centering
\captionsetup{type=table}
\captionof{table}{Complete pedestrian-network-adherence results. Values are percentages of executed action duration.}
\label{tab:full-pna}
\scriptsize
\setlength{\tabcolsep}{2.7pt}
\renewcommand{\arraystretch}{1.07}
\resizebox{\textwidth}{!}{%
\begin{tabular}{l*{13}{c}}
\toprule
\rowcolor{ModelHeadLight}
Model & VR & OU & AEQ & Short & Long & Instr. & Constr. & Search & Intent & Time & Multi & Closure & Ped. \\
\midrule
GPT-5.6-Luna & 64.7 & 71.6 & 92.2 & \underline{97.7} & 95.3 & \textbf{100.0} & \underline{99.6} & \textbf{99.8} & \underline{99.4} & 86.2 & \textbf{97.6} & \textbf{98.4} & \textbf{98.6} \\
GPT-5.5 & 63.8 & 74.8 & 91.4 & 96.9 & 94.3 & 98.9 & 96.3 & \underline{99.6} & 98.6 & 88.0 & 95.7 & 94.7 & 93.8 \\
GPT-5.4 & 66.2 & \underline{75.0} & 91.1 & 96.5 & 97.0 & \textbf{100.0} & 96.8 & 98.3 & 99.2 & 82.3 & 93.9 & \underline{96.4} & \underline{98.5} \\
GPT-5.2 & 68.0 & 74.7 & 90.9 & 94.7 & 96.4 & 99.0 & 98.1 & 96.0 & \textbf{99.5} & 82.6 & 96.8 & 94.2 & 95.4 \\
Claude-Opus-5 & \underline{69.4} & \textbf{77.9} & 92.8 & 97.5 & 93.9 & \textbf{100.0} & 93.1 & \underline{99.6} & 98.9 & 85.5 & 94.6 & 93.6 & 94.4 \\
Claude-Opus-4.6 & \textbf{70.0} & \underline{75.0} & 92.5 & 97.5 & 95.2 & 99.9 & 94.6 & 97.7 & 99.0 & 91.5 & 91.8 & 93.0 & 92.9 \\
Gemini-3.8-Flash & 63.1 & 69.8 & \textbf{94.1} & \textbf{98.0} & \underline{98.8} & \textbf{100.0} & \textbf{100.0} & 98.7 & \underline{99.4} & 91.1 & 96.7 & 95.8 & 97.5 \\
Gemini-3.6-Flash & 68.9 & 74.3 & 92.5 & 97.1 & \textbf{99.4} & \textbf{100.0} & 98.5 & 99.0 & 99.2 & 92.6 & 91.9 & 95.4 & 96.0 \\
Gemini-3.1-Pro & 65.2 & 59.1 & \underline{93.0} & 95.2 & \underline{98.8} & 99.6 & 95.1 & 96.6 & 99.2 & 88.2 & 92.7 & 93.4 & 94.0 \\
Doubao-Seed-2.0-Pro & 64.0 & 73.2 & 91.1 & 96.0 & 95.9 & 99.4 & 98.0 & 97.4 & 99.0 & \underline{93.7} & 95.9 & 93.5 & 95.0 \\
GLM-5V-Turbo & 67.5 & 71.2 & 90.6 & 97.3 & 96.9 & 99.2 & 98.6 & 98.2 & 99.1 & 92.0 & 96.1 & 93.3 & 95.9 \\
Kimi-K3 & 66.2 & 67.4 & 91.2 & 96.7 & 93.7 & 99.9 & 99.3 & 99.4 & 99.1 & \textbf{94.1} & \underline{96.9} & 95.1 & 90.6 \\
\bottomrule
\end{tabular}}
\end{minipage}

\section{Matched Retrospective Checkpoint Analysis}
\label{app:longnav-checkpoint}
To quantify endpoint regression, we compare GPT-5.5's observed final-state outcomes with retrospective oracle-rescored outcomes on the same LongNav episodes.
The oracle changes no action.
It recovers substantial partial progress without improving arrival, showing that late regression contributes to, but does not fully explain, LongNav failure.

\begin{minipage}{\textwidth}
\centering
\captionsetup{type=table}
\captionof{table}{Matched LongNav endpoint and retrospective-checkpoint results. The oracle row rescores GPT-5.5 at its closest initial or post-movement checkpoint.}
\label{tab:app-longnav-checkpoint}
\small
\setlength{\tabcolsep}{5.0pt}
\renewcommand{\arraystretch}{1.10}
\resizebox{\textwidth}{!}{%
\begin{tabular}{lcccc}
\toprule
Model / condition & Success (\%) & Closer than start (\%) & $\geq$20\% reduction (\%) & Mean remaining ratio (\%) \\
\midrule
GPT-5.5 (observed)              & 0.0 & 57.5 & 33.8 & 98.9 \\
GPT-5.5 (oracle checkpoint)     & 0.0 & 98.8 & 67.5 & 67.3 \\
\bottomrule
\end{tabular}}
\end{minipage}
\vspace{1.4em}

\section{LongNav Failure-Type Analysis}
\label{app:longnav-failure-types}

\begin{figure}[t!]
\centering
\includegraphics[width=0.95\textwidth]{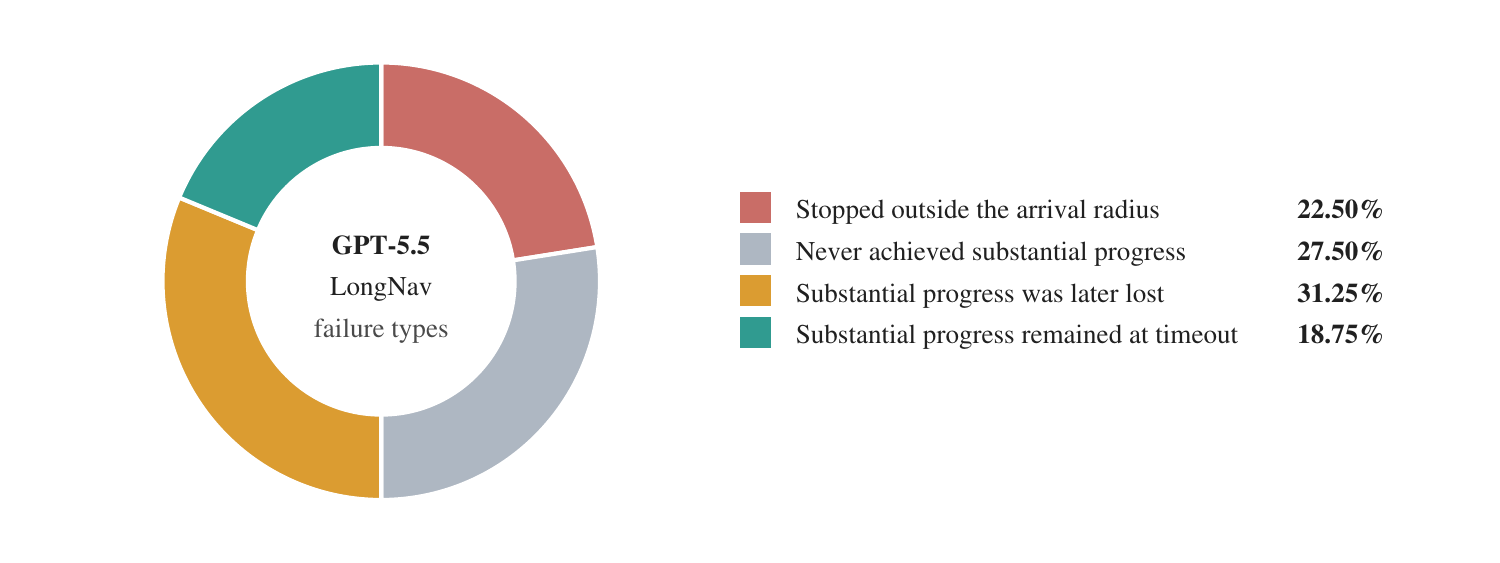}
\caption{GPT-5.5 LongNav runs grouped by their run-ending trajectory pattern. Substantial progress denotes a reduction of at least 20\% in horizontal goal distance. The final recorded checkpoint defines the endpoint of each run.}
\label{fig:app-longnav-failure-types}
\end{figure}

To characterize how MLLM agents fail on LongNav beyond endpoint success, we partition the runs into four mutually exclusive trajectory-level categories on GPT-5.5 (Figure~\ref{fig:app-longnav-failure-types}). 
\emph{Stopped outside the arrival radius} denotes cases in which the model voluntarily terminates while remaining beyond the goal tolerance. 
\emph{Never achieved substantial progress} covers runs that do not cross the predefined distance-reduction threshold at any locomotion checkpoint. 
\emph{Substantial progress was later lost} captures trajectories that temporarily approach the goal but fail to retain this improvement at the final recorded checkpoint. 
\emph{Substantial progress remained at timeout} includes runs that preserve meaningful progress at the final state but do not arrive before the interaction budget is exhausted. 
Substantial progress is defined as a reduction of at least 20\% in horizontal goal distance, and the final recorded checkpoint defines the endpoint of each run.

We find that LongNav failure is not simply an inability to initiate useful motion. 
The most prominent behavior is unstable progress, where the model approaches the goal and subsequently regresses. 
This points to weaknesses in maintaining a consistent global orientation, preserving spatial state, and recovering after a wrong turn. 
Premature stopping outside the arrival radius further indicates poorly calibrated localization or arrival verification, while progress retained until timeout suggests inefficient route execution or difficulty resolving local access constraints.

%% file: appendix_prompts.tex
%

\definecolor{PromptInk}{HTML}{25313C}
\definecolor{PromptRule}{HTML}{D8DEE4}
\definecolor{PromptVariable}{HTML}{6B7280}
\definecolor{PromptShared}{HTML}{52677D}

\lstdefinestyle{urbanpromptstyle}{
  basicstyle=\ttfamily\scriptsize\color{PromptInk},
  columns=fullflexible,
  keepspaces=true,
  showstringspaces=false,
  breaklines=true,
  breakatwhitespace=true,
  breakindent=0pt,
  breakautoindent=false,
  lineskip=1.5pt,
  emptylines=1,
  aboveskip=0pt,
  belowskip=0pt,
  xleftmargin=0pt,
  frame=none,
  literate={—}{{\textemdash}}1 {–}{{\textendash}}1 {°}{{\ensuremath{{}^{\circ}}}}1
}

\newtcblisting{urbanprompt}[3][]{
  enhanced,
  breakable,
  listing only,
  listing engine=listings,
  listing options={style=urbanpromptstyle},
  colback=#2!4!white,
  colframe=#2!72!black,
  boxrule=0.45pt,
  arc=1.5mm,
  left=1.7mm,
  right=1.7mm,
  top=1.4mm,
  bottom=1.4mm,
  before skip=6pt,
  after skip=10pt,
  title={\sffamily\bfseries\footnotesize #3},
  colbacktitle=#2!13!white,
  coltitle=#2!70!black,
  fonttitle=\sffamily\bfseries\footnotesize,
  attach boxed title to top left={xshift=1.6mm,yshift*=-1.1mm},
  boxed title style={
    boxrule=0pt,
    arc=1.2mm,
    left=1.2mm,
    right=1.2mm,
    top=0.7mm,
    bottom=0.7mm,
  },
  #1
}

\newcommand{\prompttask}[4]{%
  \Needspace{7\baselineskip}%
  \par\medskip
  \noindent
  \colorbox{#1Light}{%
    \parbox{\dimexpr\linewidth-2\fboxsep\relax}{%
      \sffamily\small
      \textcolor{#1}{\textbf{#2}}\hfill
      \textcolor{PromptInk}{#4}\enspace
      \textcolor{#1}{\textbf{(#3)}}%
    }%
  }%
  \par\smallskip
}

\section{Prompts Used in the Experiments}
\label{app:prompts}

For reproducibility, we report the task-specific prompts used by the evaluation code. Text enclosed in angle brackets denotes task-dependent content inserted at runtime. The wording outside these fields is reproduced without modification.

The public \UrbanGroundName{} API exposes additional controls for environment development. These controls were not available to the evaluated models. The evaluation runner accepted only the actions listed below and rejected every other command before execution. In particular, the models could not call \texttt{navigate}, \texttt{clear\_route}, \texttt{identify\_location}, or \texttt{map\_teleport}. The interactive map displayed the overhead scene and the markers made visible by each task. It did not compute or highlight a route. No next waypoint, shortest-path length, or remaining-distance signal was shown to the model.

\subsection{Shared Agent Interaction Protocol}

For every episode, the system message concatenates the task-specific system prompt with an action-space prompt and a ReAct output protocol. The following prompts specify how the agent learns the available actions and the required response format. Question-answering and navigation tasks use separate variants because their termination and scoring procedures differ.

\begin{urbanprompt}{PromptShared}{Question-answering action-space prompt}
Action space (choose exactly one action per exploration turn).

Every action uses one flat JSON object. Its required `action` field is one of the literal action names listed below; include only the parameters defined for that action. The descriptions below are schemas, not example actions.

First-person mode actions:
- move: `action` = "move"; `dir` is one of "forward", "backward", "left", or "right"; `seconds` is a number in [0.05, 2.0]. Optional: `yaw_rate` and `pitch_rate` are numbers in [-180, 180] degrees per second; `jump` is a boolean; `jump_at` is a number in [0, seconds].
- sprint: same fields and ranges as move, with `action` = "sprint".
- look: `action` = "look"; `yaw` is a number in [-180, 180] degrees, where positive turns right and negative turns left; `pitch` is a number in [-90, 90] degrees, where positive looks up and negative looks down.
- jump: only the `action` field with value "jump".
- open_map: only the `action` field with value "open_map".

Map mode actions:
- map_select: `action` = "map_select"; `x` and `y` are normalized screen coordinates in [0, 1], with x increasing left-to-right and y increasing top-to-bottom.
- map_pan: `action` = "map_pan"; `east` and `north` are distances in meters, each in [-2000, 2000].
- map_zoom: `action` = "map_zoom"; `factor` is a number in [0.25, 4.0]; values below 1 move closer and values above 1 move farther.
- map_orbit: `action` = "map_orbit"; `yaw` is a number in [-180, 180] degrees and `pitch` is a number in [-90, 90] degrees.
- close_map: only the `action` field with value "close_map".

Available in both first-person and map mode:
- terminate: only the `action` field with value "terminate". Choose it once you have enough visual evidence to answer confidently; exploration ends immediately and you will then be asked for the final multiple-choice answer.

Use first-person actions only while viewing the first-person scene and map actions only while the map is visible. Never request move or sprint for longer than 2 seconds. If a larger duration is supplied, the environment clamps it and executes only 2 seconds. Do not infer or emit actions outside this complete action space.
\end{urbanprompt}

\begin{urbanprompt}{PromptShared}{Question-answering ReAct protocol prompt}
Follow a visual ReAct loop on every exploration turn:
1. Observation: extract only relevant visible evidence from the current screenshot.
2. Reason: use that evidence and your remembered prior observations/actions to decide what to inspect next.
3. Action: choose exactly one action from the action space.

Return exactly one JSON object and no Markdown or extra text. The top-level object must contain exactly these fields:
- `observation`: a non-empty string containing concise visible evidence.
- `reason`: a non-empty string containing the concise reason for the next action.
- `action`: one flat action object conforming to exactly one schema in the action space above.

The nested action object must use the string field `action` for its action name. Do not use a `type` field, do not key the object by the action name, and do not add unavailable parameters. No concrete numeric action example is provided; select every parameter solely from current visual evidence and conversation memory.
During exploration, do not answer the multiple-choice question. When the visible evidence is sufficient, choose terminate; you will then be explicitly asked for the final answer. If you do not terminate, exploration ends automatically when the turn limit is reached.
You receive only task text, screenshots, and conversation memory. Never assume access to hidden simulator state.
\end{urbanprompt}

When exploration ends, question-answering tasks receive the following final-answer format instruction.

\begin{urbanprompt}{PromptShared}{Question-answering final-answer schema}
Return exactly one JSON object with no Markdown or extra text:
{"answer":"A|B|C|D","reason":"brief evidence-based reason grounded in the visual exploration"}
\end{urbanprompt}

\begin{urbanprompt}{PromptShared}{Navigation action-space prompt}
Action space (choose exactly one action per navigation turn).

Every action uses one flat JSON object. Its required `action` field is one of the literal action names listed below; include only the parameters defined for that action. The descriptions below are schemas, not example actions.

First-person mode actions:
- move: `action` = "move"; `dir` is one of "forward", "backward", "left", or "right"; `seconds` is a number in [0.05, 2.0]. Optional: `yaw_rate` and `pitch_rate` are numbers in [-180, 180] degrees per second; `jump` is a boolean; `jump_at` is a number in [0, seconds].
- sprint: same fields and ranges as move, with `action` = "sprint".
- look: `action` = "look"; `yaw` is a number in [-180, 180] degrees, where positive turns right and negative turns left; `pitch` is a number in [-90, 90] degrees, where positive looks up and negative looks down.
- jump: only the `action` field with value "jump".
- open_map: only the `action` field with value "open_map".

Map mode actions:
- map_select: `action` = "map_select"; `x` and `y` are normalized screen coordinates in [0, 1], with x increasing left-to-right and y increasing top-to-bottom. Selecting a point changes the visible map selection. It does not invoke location lookup or route computation.
- map_pan: `action` = "map_pan"; `east` and `north` are distances in meters, each in [-2000, 2000].
- map_zoom: `action` = "map_zoom"; `factor` is a number in [0.25, 4.0]; values below 1 move closer and values above 1 move farther.
- map_orbit: `action` = "map_orbit"; `yaw` is a number in [-180, 180] degrees and `pitch` is a number in [-90, 90] degrees.
- close_map: only the `action` field with value "close_map".

The actions listed above form the complete model-facing map interface. No route-computation action is available.

Available in both first-person and map mode:
- terminate: only the `action` field with value "terminate". Choose it when you believe the task is complete or deliberately want to stop; navigation ends immediately and your current position/state is scored.

Use first-person actions only while viewing the first-person scene and map actions only while the map is visible. Never request move or sprint for longer than 2 seconds. If a larger duration is supplied, the environment clamps it and executes only 2 seconds. Do not infer or emit actions outside this complete action space.
\end{urbanprompt}

\begin{urbanprompt}{PromptShared}{Navigation ReAct protocol prompt}
Follow a visual ReAct loop on every navigation turn:
1. Observation: extract only relevant visible evidence from the current screenshot (street layout, signs, crossings, obstacles).
2. Reason: use that evidence and your remembered prior observations/actions to decide what to do next in order to reach the destination.
3. Action: choose exactly one action from the action space.

Return exactly one JSON object and no Markdown or extra text. The top-level object must contain exactly these fields:
- `observation`: a non-empty string containing concise visible evidence.
- `reason`: a non-empty string containing the concise reason for the next action.
- `action`: one flat action object conforming to exactly one schema in the action space above.

The nested action object must use the string field `action` for its action name. Do not use a `type` field, do not key the object by the action name, and do not add unavailable parameters. No concrete numeric action example is provided; select every parameter solely from current visual evidence and conversation memory.
Keep navigating turn after turn until you believe you have arrived at the destination, then choose terminate so the current state can be scored. You may also choose terminate if you deliberately decide to stop. If you do not terminate, navigation ends automatically on arrival or when the turn limit is reached; there is no final answer to submit for this task.
You receive only task text, screenshots, and conversation memory. A fixed start or goal description may be included in the task instruction. Never assume access to updated simulator state such as current coordinates or a distance-remaining readout. If you need global context, open the map and inspect the markers made visible by the task. The map does not compute or display a route.
\end{urbanprompt}

\subsection{RQ1 / Level 1: Local Environment Understanding}

Visual Recognition (VR), Orientation Understanding (OU), and Active Exploration Questions (AEQ) share the following multiple-choice task format.

\begin{urbanprompt}{LevelOne}{Runtime task description for VR, OU, and AEQ}
<QUESTION>
A. <OPTION A>
B. <OPTION B>
C. <OPTION C>
D. <OPTION D>
\end{urbanprompt}

\prompttask{LevelOne}{Level 1}{VR}{Visual Recognition}

\begin{urbanprompt}{LevelOne}{System prompt}
You are solving a landmark-recognition multiple-choice task in a photorealistic Hong Kong simulation.

Keep the question and choices in conversation memory and actively gather visual evidence relevant to distinguishing them. Inspect storefronts, signs, buildings, objects, colors, and nearby facilities. Avoid aimless travel and repeated views; prefer camera turns and short movements that improve visibility of relevant landmarks or text.

Do not answer until explicitly asked for the final answer. You will not receive hidden simulator coordinates, orientation angles, road metadata, or other privileged state.
\end{urbanprompt}

\prompttask{LevelOne}{Level 1}{OU}{Orientation Understanding}

\begin{urbanprompt}{LevelOne}{System prompt}
You are solving an orientation-understanding multiple-choice task in a photorealistic Hong Kong simulation.

Preserve the original visible starting orientation as the reference frame. Gather visual evidence about whether the target is in front, behind, left, or right, or about its compass/facing direction. Remember your own relative camera-turn and movement actions so later observations are not confused with the initial view.

Prefer controlled turns and short movements; avoid unnecessary displacement that destroys the useful spatial reference. Do not answer until explicitly asked for the final answer.

You will not receive hidden simulator coordinates, absolute compass angles, road metadata, or other privileged state.
\end{urbanprompt}

\prompttask{LevelOne}{Level 1}{AEQ}{Active Exploration Questions}

\begin{urbanprompt}{LevelOne}{System prompt}
You are solving an active-search multiple-choice task in a photorealistic Hong Kong simulation.

The answer may not be visible from the initial viewpoint. Actively explore the nearby environment to find direct visual evidence: first scan around with controlled camera turns, then move along accessible nearby streets when necessary, inspect street signs, storefronts, building names, public-facility signs, and other relevant landmarks.

Use the question and choices to guide a systematic local search. Avoid standing still, repeatedly viewing the same scene, or assuming an answer from general geographic knowledge. Prefer direct evidence visible in screenshots. Keep a useful memory of which directions and paths you have already inspected so exploration covers new nearby areas.

Do not answer during exploration; continue gathering evidence until explicitly asked for the final answer. You will not receive hidden simulator coordinates, absolute orientation angles, road metadata, search results, or other privileged state.
\end{urbanprompt}

\subsection{RQ2 / Level 2: Navigation under Explicit Instructions}

\prompttask{LevelTwo}{Level 2}{SGN}{Short-range Goal Navigation}

\begin{urbanprompt}{LevelTwo}{System prompt}
You are solving a short-range navigation task in a photorealistic Hong Kong simulation.

The destination is within visual range of the starting point. Use whichever combination of first-person movement, turning, and map actions you find most effective, and stay on sidewalks and other pedestrian infrastructure whenever possible instead of cutting through roads or private property.

Continue moving turn after turn until you judge that you have arrived at the destination or you run out of turns. There is no multiple-choice answer to submit for this task; your only goal is to physically reach the destination described in the task text.

You will not receive hidden simulator coordinates, a distance-remaining readout, or other privileged state; rely only on what is visible in each screenshot and your own memory of the route travelled so far.
\end{urbanprompt}

\begin{urbanprompt}{LevelTwo}{Runtime task instruction}
[Short Navigation Task]
Start: <START LOCATION>
Goal:  <GOAL LOCATION>

The start and end points are within visual range, so navigating purely by observation is usually fastest, but you may inspect the map if you find it helpful. No computed route is available.
\end{urbanprompt}

\prompttask{LevelTwo}{Level 2}{LGN}{Long-range Goal Navigation}

\begin{urbanprompt}{LevelTwo}{System prompt}
You are solving a long-range navigation task in a photorealistic Hong Kong simulation.

The destination is far from the starting point and is unlikely to be visible directly. The interactive map shows your current position and the task destination. You may use map actions to inspect their spatial relation, but the map does not compute or highlight a route. Determine the route yourself from the map and first-person observations. Prefer sidewalks, footbridges, subways, and other pedestrian infrastructure over cutting through roads or private property.

Continue navigating turn after turn until you judge that you have arrived at the destination or you run out of turns. There is no multiple-choice answer to submit for this task; your only goal is to physically reach the destination described in the task text.

The task instruction provides fixed start and goal descriptions. You will not receive updated coordinates, a distance-remaining readout, or any route guidance. Rely only on what is visible in each screenshot and your own memory of previous observations and actions.
\end{urbanprompt}

\begin{urbanprompt}{LevelTwo}{Runtime task instruction}
[Long-Range Navigation Task]
Start: <START LOCATION>
Goal:  <GOAL LOCATION>

The destination is far away. Use first-person observations and the interactive map to determine and execute a route. No computed or highlighted route is available.
\end{urbanprompt}

\prompttask{LevelTwo}{Level 2}{IN}{Instructional Navigation}

\begin{urbanprompt}{LevelTwo}{System prompt}
You are solving an instruction-following navigation task in a photorealistic Hong Kong simulation.

You are given a sequence of turn-by-turn natural-language instructions (for example: "go straight to the fork ahead and turn left, then continue to the next fork and turn right"). Your goal is to execute these instructions in order, identifying forks, junctions, crossings, and landmarks mentioned in the text from what you see in each screenshot, and stay on sidewalks and other pedestrian infrastructure whenever possible instead of cutting through roads or private property.

Continue moving turn after turn until you judge that you have completed the instructions and arrived at the implied destination, or you run out of turns. There is no multiple-choice answer to submit for this task; your only goal is to physically follow the instructions to their end point.

You will not receive hidden simulator coordinates, a distance-remaining readout, or other privileged state; rely only on what is visible in each screenshot, the instruction text, and your own memory of the route travelled and instructions already completed so far.
\end{urbanprompt}

\begin{urbanprompt}{LevelTwo}{Runtime task instruction}
[Instruction-Following Navigation Task]

Turn-by-turn instructions:
<TURN-BY-TURN INSTRUCTIONS>
\end{urbanprompt}

\prompttask{LevelTwo}{Level 2}{CN}{Constrained Navigation}

\begin{urbanprompt}{LevelTwo}{System prompt}
You are solving a constrained navigation task in a photorealistic Hong Kong simulation.

One or more road segments are closed for the entire task, from the very first turn. Each closed segment is described as a short chain of waypoints; the closure itself is the line connecting consecutive waypoints in that chain, not an enclosed area. You must never cross any of these closure lines with your own movement, in either direction, at any point during the episode -- crossing one at any time immediately fails the task, even if you would otherwise reach the destination.

Before moving, open the map to see exactly where the closed segments are relative to your position and the destination, and plan a route that goes around them. Use whichever combination of first-person movement, turning, and map actions you find most effective, and stay on sidewalks and other pedestrian infrastructure whenever possible instead of cutting through roads or private property.

Continue moving turn after turn until you judge that you have arrived at the destination without ever crossing a closed segment, or you run out of turns. There is no multiple-choice answer to submit for this task; your only goal is to physically reach the destination while respecting every closure for the whole episode.

You will not receive hidden simulator coordinates, a distance-remaining readout, or an automatic crossing warning; rely only on what is visible in each screenshot (including the map), the closure descriptions given in the task text, and your own memory of the route travelled so far.
\end{urbanprompt}

\begin{urbanprompt}{LevelTwo}{Runtime task instruction}
[Constrained Navigation Task]
Start: <START LOCATION>
Goal:  <GOAL LOCATION>

<TASK DESCRIPTION>

Road closures in effect for the entire task (never cross any of these lines):
<ROAD CLOSURES>

Open the map before setting off to see these closures relative to your position and the destination, and plan a route around them from the very first step.
\end{urbanprompt}

\subsection{RQ2 / Level 3: Exploration under Implicit Instructions}

\prompttask{LevelThree}{Level 3}{PTS}{Place-type Search}

\begin{urbanprompt}{LevelThree}{System prompt}
You are solving a place-type search task in a photorealistic Hong Kong simulation.

You are asked to take the user to a nearby place of a requested type or name (for example "Go to the nearby park", "Find the nearest public toilet", "Take me to City Hall"). You must figure out where such a place is and physically travel to it: use the map to locate candidate facilities around you, inspect street-level signage and storefronts to confirm what a place is, and navigate there.

Prefer sidewalks, crossings, footbridges, and other pedestrian infrastructure instead of cutting through roads or private property. Keep track of the streets you have already searched so you do not wander in circles.

Once you believe you have arrived at the requested place, stop moving and take a clear look at its entrance or signage: your final position and view are what the judge will see. There is no multiple-choice answer to submit; success is decided by whether your final position counts as having arrived at the requested place.

You will not receive hidden simulator coordinates, a distance-remaining readout, search results, or other privileged state; rely only on what is visible in each screenshot and your own memory of the route travelled so far.
\end{urbanprompt}

\begin{urbanprompt}{LevelThree}{Runtime task instruction}
[Place-Type Search Task]
Current location: <CURRENT LOCATION>

Request: <PLACE REQUEST>

Locate a suitable nearby place that satisfies the request, navigate to it, and stop at its entrance or in front of its signage. You may use the map, visual inspection, and first-person movement in any combination.
\end{urbanprompt}

The following task-specific reminder is inserted before each observation.

\begin{urbanprompt}{LevelThree}{Per-turn context}
Reminder: your goal is to reach the requested place (<PLACE REQUEST>). If you believe you have arrived, stop there and make sure the place or its signage is clearly visible.
\end{urbanprompt}

\prompttask{LevelThree}{Level 3}{III}{Implicit Intent Inference}

\begin{urbanprompt}{LevelThree}{System prompt}
You are solving an implicit-intent navigation task in a photorealistic Hong Kong simulation.

You are given a short, everyday-life goal that does not directly name the destination (for example: "Go deposit some money", "My eyes are acting up — I need a professional eye check", or "Where can I buy watercolor supplies?"). First infer the kind of real-world place that would satisfy this goal (bank/ATM, optician/eye clinic, art supply store, pharmacy, mobile carrier shop, etc.). Then find and physically travel to a suitable nearby POI of that type using only what you can observe.

Use controlled camera turns, first-person movement, and the map strategically. Inspect storefront names, signs, logos, entrances, and other direct visual evidence before deciding that a place satisfies the intent.

Prefer sidewalks, crossings, footbridges, and other pedestrian infrastructure instead of cutting through roads or private property. Keep a memory of streets already searched so exploration covers new nearby areas.

Continue moving until you judge that you have reached an appropriate destination. There is no multiple-choice answer to submit for this task. You will not receive hidden simulator coordinates, a distance-remaining readout, search results, or other privileged state.
\end{urbanprompt}

\begin{urbanprompt}{LevelThree}{Runtime task instruction}
[Implicit Intent Navigation Task]
Current location: <CURRENT LOCATION>

Everyday goal: <EVERYDAY GOAL>

Infer the type of place that would satisfy this goal, find a suitable nearby POI from visible signs and map evidence, and navigate to it. The destination category is intentionally not stated directly. You may use the map, visual inspection, and first-person movement in any combination.
\end{urbanprompt}

\subsection{RQ2 / Level 4: Multi-Task Planning}

\prompttask{LevelFour}{Level 4}{TWS}{Time-window Scheduling}

\begin{urbanprompt}{LevelFour}{System prompt}
You are solving a time-window schedule-following task in a photorealistic Hong Kong simulation.

You are given your schedule for the day: an ordered list of appointments at different places, each with a scheduled time (a time window you must respect). Your job is to physically travel to every appointment place, in the scheduled order, and to arrive at each one before its scheduled time passes. Being late is a failure even if you eventually arrive, and visiting places out of order is a failure even if every place is reached.

Please plan an efficient route, prefer direct movement toward the next appointment, and avoid unnecessary detours or idle inspection. At the start of every turn you are told the current time, which appointments you have already completed, and how much time remains until the next appointment. Use that information to decide when to hurry.

Use controlled camera turns, first-person movement, and the map strategically. The next appointment's destination is marked on the map; open the map whenever you need to re-orient. Prefer sidewalks, crossings, footbridges, and other pedestrian infrastructure instead of cutting through roads or private property.

Once you arrive at an appointment place, move on to the next appointment immediately. There is no multiple-choice answer to submit for this task; success is measured by whether you reach every appointment place, in order, and on time. You will not receive hidden simulator coordinates or a distance-remaining readout.
\end{urbanprompt}

\begin{urbanprompt}{LevelFour}{Runtime task instruction}
[Time-Window Schedule Task]
Current location: <CURRENT LOCATION>

It is currently <CURRENT TIME>.
Your appointments today, in the order you must visit them:
1. <PLACE> — <ACTIVITY> — due by <DEADLINE>
...
Everything must be finished by <OVERALL DEADLINE>.

Visit the appointments in this exact order and arrive before each scheduled time. The clock keeps running while you act.
\end{urbanprompt}

The following status is inserted before each observation, with unavailable clauses omitted.

\begin{urbanprompt}{LevelFour}{Per-turn context}
Schedule status: <COMPLETED>/<TOTAL> appointments completed.
Current time: <CURRENT TIME>. Schedule status: <COMPLETED>/<TOTAL> appointments completed.
Next appointment: <PLACE> (<ACTIVITY>), due <DEADLINE> — <MINUTES LEFT> minutes left.
— OVERDUE by <MINUTES OVERDUE> minutes.
All appointments completed.
\end{urbanprompt}

\prompttask{LevelFour}{Level 4}{MSP}{Multi-stop Route Planning}

\begin{urbanprompt}{LevelFour}{System prompt}
You are solving a multi-point route-planning task in a photorealistic Hong Kong simulation.

You must visit several destination places in one outing. There is NO required visiting order and no time schedule: the whole point of the task is to plan an efficient route yourself. Before setting off, think about where each destination is relative to you and to each other (open the map: every destination is marked on it), pick a visiting order that minimizes backtracking, and then follow your plan.

Use controlled camera turns, first-person movement, and the map strategically. Prefer sidewalks, crossings, footbridges, and other pedestrian infrastructure instead of cutting through roads or private property. When you believe you have arrived at one destination, move on to the next unvisited one immediately; do not linger.

The task ends when you have visited every destination or you run out of turns. There is no multiple-choice answer to submit; success is measured by how many of the destinations you physically reach and how efficient your route is. You will not receive hidden simulator coordinates or a distance-remaining readout.
\end{urbanprompt}

\begin{urbanprompt}{LevelFour}{Runtime task instruction}
[Multi-Point Route Planning Task]
Current location: <CURRENT LOCATION>

<TASK DESCRIPTION>

There are <NUMBER OF DESTINATIONS> destinations to visit. There is no required order: plan the most efficient route yourself, then visit all of them.
\end{urbanprompt}

\begin{urbanprompt}{LevelFour}{Per-turn context}
Route progress: <VISITED>/<TOTAL> destinations visited. Remaining: <REMAINING DESTINATIONS>.
\end{urbanprompt}

\subsection{RQ3 / Level 5: Dynamic Environment Interaction}

\prompttask{LevelFive}{Level 5}{DCR}{Dynamic Road-closure Replanning}

\begin{urbanprompt}{LevelFive}{System prompt}
You are solving a navigation task in a photorealistic Hong Kong simulation.

At the start, navigate exactly as you would for an ordinary point-to-point trip: there is no known obstruction and you should head toward the destination by whichever combination of first-person movement, turning, and map actions you find most effective, staying on sidewalks and other pedestrian infrastructure whenever possible instead of cutting through roads or private property.

Partway through the episode, a road closure may newly appear on the map -- you will be told explicitly, in a system notice inside your observation, exactly when this happens. From the moment you receive that notice onward, one or more road segments become closed for the remainder of the episode. Each closed segment is described as a short chain of waypoints; the closure itself is the line connecting consecutive waypoints in that chain, not an enclosed area. Once notified, you must never cross any of these closure lines with your own movement, in either direction, for the rest of the episode -- crossing one after being notified immediately fails the task, even if you would otherwise reach the destination. Crossing the same ground before you are notified is not a violation, since the closure did not exist yet from your perspective.

As soon as you are notified of a closure, open the map to see exactly where it is relative to your current position and the destination, and replan a route that goes around it.

Continue moving turn after turn until you judge that you have arrived at the destination without crossing a closure after being notified of it, or you run out of turns. There is no multiple-choice answer to submit for this task; your only goal is to physically reach the destination while respecting any closure disclosed to you.

You will not receive hidden simulator coordinates, a distance-remaining readout, or an automatic crossing warning; rely only on what is visible in each screenshot (including the map), the closure notice once given, and your own memory of the route travelled so far.
\end{urbanprompt}

\begin{urbanprompt}{LevelFive}{Runtime task instruction before disclosure}
[Navigation Task]
Start: <START LOCATION>
Goal:  <GOAL LOCATION>

<TASK DESCRIPTION>

Navigate to the destination. There is no known obstruction right now; use whichever combination of first-person movement, turning, and map actions you find most effective.
\end{urbanprompt}

After the closure appears, the following notice is inserted once before the next observation.

\begin{urbanprompt}{LevelFive}{One-off closure notice}
NEW SYSTEM NOTICE: A road closure has just appeared on the map. The following segment(s) are now closed and must not be crossed for the remainder of this task (crossing was not a violation before this notice, but is a violation from now on):
<ROAD CLOSURES>

Open the map now to see exactly where this closure is relative to your current position and the destination, and replan your route around it.
\end{urbanprompt}

\prompttask{LevelFive}{Level 5}{NP}{Navigation among Pedestrians}

Navigation among Pedestrians (NP) uses the same system prompt and task instruction as Long-range Goal Navigation (LGN).

\begin{urbanprompt}{LevelFive}{System prompt}
You are solving a long-range navigation task in a photorealistic Hong Kong simulation.

The destination is far from the starting point and is unlikely to be visible directly. The interactive map shows your current position and the task destination. You may use map actions to inspect their spatial relation, but the map does not compute or highlight a route. Determine the route yourself from the map and first-person observations. Prefer sidewalks, footbridges, subways, and other pedestrian infrastructure over cutting through roads or private property.

Continue navigating turn after turn until you judge that you have arrived at the destination or you run out of turns. There is no multiple-choice answer to submit for this task; your only goal is to physically reach the destination described in the task text.

The task instruction provides fixed start and goal descriptions. You will not receive updated coordinates, a distance-remaining readout, or any route guidance. Rely only on what is visible in each screenshot and your own memory of previous observations and actions.
\end{urbanprompt}

\begin{urbanprompt}{LevelFive}{Runtime task instruction}
[Long-Range Navigation Task]
Start: <START LOCATION>
Goal:  <GOAL LOCATION>

The destination is far away. Use first-person observations and the interactive map to determine and execute a route. No computed or highlighted route is available.
\end{urbanprompt}